\documentclass[12pt]{article}
\usepackage{amsmath}
\usepackage{graphicx}
\usepackage{natbib}
\usepackage{url} 

\newcommand{\blind}{0}

\usepackage{pstricks, pst-node, psfrag}
\usepackage{amssymb,amsmath}
\usepackage{verbatim,enumerate}
\usepackage{rotating, lscape}
\usepackage{setspace}
\usepackage{algorithm}
\usepackage{algpseudocode}
\algrenewcommand\algorithmicrequire{\textbf{Input:}}
\algrenewcommand\algorithmicensure{\textbf{Output:}}

\usepackage{bbm}
\usepackage{natbib}
\usepackage{amsthm}
\usepackage{tikz}
\usepackage{etoolbox} 
\usepackage{listofitems} 
\usepackage{url}
\usepackage{appendix}
\usepackage{graphicx}
\graphicspath{ {./Figures/} }
\usepackage{bm}
\usepackage{color}
\usepackage[colorlinks=true,
            linkcolor=darkgray,
            urlcolor=blue,
            citecolor=brown]{hyperref}
\usepackage{longtable}
\usepackage{calc}
\usepackage{float}
\usepackage{multirow}
\usepackage{arydshln}
\usepackage{enumitem}
\usepackage{eucal}
\usepackage{subcaption}
\usepackage{amsfonts,amsmath,amssymb,amsthm}
\usepackage{booktabs}

\usepackage{lipsum}
\tikzset{>=latex} 
\colorlet{myred}{red!80!black}
\colorlet{myblue}{blue!80!black}
\colorlet{mygreen}{green!60!black}
\colorlet{mydarkred}{myred!40!black}
\colorlet{mydarkblue}{myblue!40!black}
\colorlet{mydarkgreen}{mygreen!40!black}
\tikzstyle{node}=[very thick,circle,draw=myblue,minimum size=22,inner sep=0.5,outer sep=0.6]
\tikzstyle{connect}=[->,thick,mydarkblue,shorten >=1]
\tikzset{ 
	node 1/.style={node,mydarkgreen,draw=mygreen,fill=mygreen!25},
	node 2/.style={node,mydarkblue,draw=myblue,fill=myblue!20},
	node 3/.style={node,mydarkred,draw=myred,fill=myred!20},
}

\renewcommand{\baselinestretch}{1.5} 
\theoremstyle{definition}

\newcommand{\bi}{\begin{itemize}}
\newcommand{\ei}{\end{itemize}}
\newcommand{\pkg}[1]{\texttt{#1}}
\newcommand{\bfs}{\mathbf{s}}

\begin{document}

\def\spacingset#1{\renewcommand{\baselinestretch}%
{#1}\small\normalsize} \spacingset{1}

 
\if0\blind
{
  \title{\bf Efficient Large-scale Nonstationary Spatial Covariance Function Estimation Using Convolutional Neural Networks}
  \author{Pratik Nag, Yiping Hong, Sameh Abdulah, Ghulam A. Qadir, \\Marc G. Genton, and Ying Sun \hspace{.2cm}\\
}
  \maketitle
} \fi

\if1\blind
{
  \bigskip
  \bigskip
  \bigskip
  \begin{center}
    {\LARGE\bf Title}
\end{center}
  \medskip
} \fi

\begin{abstract}
Spatial processes observed in various fields, such as climate and environmental science, often occur on a large scale and demonstrate spatial nonstationarity. Fitting a Gaussian process with a nonstationary Mat\'ern covariance is challenging. Previous studies in the literature have tackled this challenge by employing spatial partitioning techniques to estimate the parameters that vary spatially in the covariance function. The selection of partitions is an important consideration, but it is often subjective and lacks a data-driven approach.   To address this issue, in this study, we utilize the power of Convolutional Neural Networks (ConvNets) to derive subregions from the nonstationary data. We employ a selection mechanism to identify subregions that exhibit similar behavior to stationary fields. In order to distinguish between stationary and nonstationary random fields, we conducted training on ConvNet using various simulated data. These simulations are generated from Gaussian processes with Mat\'ern covariance models under a wide range of parameter settings, ensuring adequate representation of both stationary and nonstationary spatial data. We assess the performance of the proposed method with synthetic and real datasets at a large scale. The results revealed enhanced accuracy in parameter estimations when relying on ConvNet-based partition compared to traditional user-defined approaches.

\end{abstract}

\noindent%
{\it Keywords:}  Convolutional neural networks, Geospatial data, High Performance Computing, Likelihood, Nonstationary Mat\'ern Covariance, Spatial domain partitions.
\vfill

\newpage
\spacingset{1.5} 
\section{Introduction}

Gaussian processes (GPs) are statistical models that are widely used in various fields, including machine learning, statistics, signal processing, and physics. They are flexible and powerful tools for modeling complex systems, making predictions and quantifying uncertainty. In spatial statistics, GPs are widely employed for modeling, simulation and prediction (kriging) of spatial and spatio-temporal data. The spatial data modeling in  the context of GPs frequently incorporates the covariance stationarity assumption which imposes translation invariance spatial dependence properties over the entire spatial domain of interest. Stationary GPs are commonly applied for modeling and predicting various spatial processes, including but not limited to temperature, rainfall, and air pollution. However, spatial datasets often exhibit nonstationarity, which means that the statistical properties of the data vary across different parts of the spatial domain. The prevalence of nonstationarity can be attributed to several factors, including the spatial evolution in the underlying physical processes, heterogeneity in terrain between land and the ocean, variations in topographical elevation, and other related factors. In this study, we assume the mean function is constant and focus on the covariance nonstationarity.

\cite{risser2016review} reviewed various approaches to modeling covariance nonstationarity. The spatial deformation method is a well-known technique for modeling nonstationary spatial data \citep{sampson1992nonparametric, anderes2008estimating, qadir2021estimation}. It is often coupled with GPs to introduce nonstationarity into the model. In deformation-based nonstationary GPs, the estimation process involves finding an injective deformation function, that transforms the spatial coordinates. This transformation aims to make the process stationary when viewed in the transformed space of coordinates, often referred to as the deformed space. While the deformation-based approach offers flexibility for modeling nonstationary spatial data, its success heavily relies on accurately specifying and estimating the deformation function, which can be challenging when dealing with a single realization of the process. Additionally, the deformation-based method can be computationally intensive, particularly for large datasets or high-dimensional problems. The differential operator approach by \cite{jun2008nonstationary} and \cite{lindgren2011explicit} is an another alternative method for nonstationary spatial data modeling, which involves the use of differential operators to capture the spatial or temporal variation in the data. This approach allows for flexible modeling of nonstationarity and can capture a wide range of spatial or temporal patterns. Although it can be extended to big datasets, the computation can become intensive. Another well-established alternate method for nonstationary spatial modeling is the process convolution approach \citep{paciorek2006spatial,wilson2013gaussian}. This approach is based on the convolution of a stationary process with a nonstationary kernel to obtain a nonstationary process. However, scalability of the the process convolution method for large datasets is still a computationally challenging task.  

Among the nonstationary methods mentioned, the convolution-based approach is arguably a preferred choice among geostatisticians. This approach has resulted in a closed-form nonstationary Mat\'ern covariance model, adding to its appeal and practicality in nonstationary spatial data modeling. The nonstationary Mat\'ern covariance model allows for spatially varying covariance parameters which characterize different types of nonstationarity. Accordingly, the parameters can be expressed as a function of spatial coordinates that vary across space. This spatially varying specification of covariance parameters poses challenges in estimation, as the number of estimable parameters increases with the growing number of spatial locations. Accuracy and efficiency in such an estimation are challenging to achieve. To address this problem, many implementations assume that the covariance function is locally stationary \citep{paciorek2006spatial,anderes2011local} or weighted locally stationary \citep{risser2015local,fouedjio2016generalized,li2019efficient}. These assumptions are rooted in the partitioning of the spatial domain into distinct subregions, where the spatially varying covariance parameters are assumed to be either constant or exhibit slow changes within each subregion. This simplification greatly facilitates model fitting in numerous applications. However, as a result, the choice of region partitioning has a substantial impact on the model's performance. Presently, the existing literature primarily relies on subjective partitioning methods or fixed splitting methods, thus lacking a data-driven approach.
 
In an alternative perspective, spatial data can be conceptualized as an incomplete image, wherein certain pixels are missing or deleted. In contrast to traditional images that use RGB values to describe each pixel, spatial data are represented by continuous values that convey specific information related to the spatial domain. As a result, the analysis of spatial data can be viewed as a specialized analog of image processing.  In image processing, deep learning models which are composed of multiple layers of nonlinear processing, have shown impressive results in feature extraction, transformation, pattern analysis, and classification. Convolutional neural networks (ConvNets), first introduced by \cite{lecun1989backpropagation}, have emerged as the leading architecture for image recognition, classification, and detection tasks \citep{lecun2015deep, zhang2016joint}. Interestingly, ConvNets are increasingly making their presence felt in the field of spatial statistics, as evidenced by recent studies \citep{gerber2021fast,lenzi2023neural,lenzi2023towards}. In line with this, we present a novel data-driven framework that leverages the image processing capabilites of the modern ConvNets to identify optimal subregions for nonstationary spatial data modeling. The optimal choice utilizes the classification of spatial fields through ConvNets into stationary and nonstationary spatial fields. This framework is based on analyzing the nonstationarity index which characterizes the prevalent nonstationarity observed in spatial data. Once the optimal subregions are identified, we estimate the parameters for each subregion assuming the stationary Matérn covariance model. To achieve scalable parameter estimation, we employ the \pkg{ExaGeoStat} framework, which is specifically designed for modeling large-scale geospatial data \citep{abdulah2018ExaGeoStat}. By utilizing this framework, we enhance the estimation of spatially varying parameters for nonstationary covariance functions, surpassing the limitations of traditional fixed-splitting approaches.

The paper is structured as follows: Section~\ref{Sec:Preliminaries} presents an introduction to the nonstationary Mat\'ern covariance function and its implementation procedure in \pkg{ExaGeoStat} \citep{abdulah2018ExaGeoStat}. Section~\ref{Sec:ConvNet_model} discusses the ConvNet algorithm, including its implementation for both stationary and nonstationary classification. Subsection~\ref{Sec:Data_Preprocessing} describes the data pre-processing which is required for reconstructing irregularly spaced data into a $100 \times 100$ gridded input for the ConvNet model. Subsection~\ref{Sec:clustering} discusses the algorithm of subregion selection through the ConvNet model. Section~\ref{Sec:Experiment} presents the empirical evaluation of the proposed approach on synthetic datasets. Section~\ref{Sec:real_data} applies the proposed approach to soil moisture data. Conclusions and discussions are provided in Section~\ref{Sec:discussion}.

\section{Gaussian Likelihood Computation} \label{Sec:Preliminaries}
\subsection{Nonstationary Mat\'ern Covariance Function}\label{nonstat_matern}

The univariate Gaussian random field (GRF), $\{Z(\bm{s}),\ \bm{s} \in D\subset \mathbb{R}^d\}$ with spatial location indexing $\bm{s}$, can be expressed as:
\begin{equation}
Z(\bm{s}) = \mu(\bm{s}) + Y(\bm{s}) + \epsilon(\bm{s}), \quad \bm{s}\in D,
\label{eq1}
\end{equation}
where $\mu(\cdot)$ is the mean function, $Y(\cdot)$ is the zero-mean GP with covariance function $C(\cdot,\cdot)$ and $\epsilon(\cdot) \sim {\cal N}(0,\tau^2(\cdot))$ is the nugget effect attributed to measurement inaccuracy and environmental variability. Here $\mu(\cdot)$ is assumed to be constant for simplicity, the covariance function $C(\cdot,\cdot) = C(\cdot,\cdot;\boldsymbol{\theta}_0)$ has a parametric form with associated parameters $\boldsymbol{\theta}_0 \in \mathbb{R}^p$, and the random processes $Y(\cdot)$ and $\epsilon(\cdot)$ are independent. The equation~\eqref{eq1} provides a generalized representation of a GRF, which accommodates both stationary and nonstationary processes.

Let $\boldsymbol{\theta} = (\boldsymbol{\theta}_0^\top, \tau^2)^\top$ represent a vector of unknown parameters. We consider $C(\cdot,\cdot) = C(\cdot,\cdot;\boldsymbol{\theta})$ to be an isotropic covariance function that incorporates the nugget term as an addition on the diagonal of the associated covariance matrix. Isotropic covariance functions exhibit rotational and reflection invariance, which is a restriction of the translation-invariant stationary covariance functions. The literature on spatial statistics offers a variety of isotropic covariance models, including the exponential, Mat\'ern, and Cauchy models \citep{cressie2015statistics}. The generalization of an isotropic covariance function into an anisotropic nonstationary covariance function is a complex and challenging task. The Mat\'ern class achieved this non-trivial generalization through the process convolution approach \citep{higdon1998process,stein2005nonstationary,paciorek2003nonstationary,paciorek2006spatial}, and as a result, we have a closed-form nonstationary Mat\'{e}rn covariance function  $C^{NS}\left(\cdot,\cdot;\boldsymbol{\theta}\right): \mathbb{R}^d\times \mathbb{R}^d\rightarrow \mathbb{R}$:
\begin{equation}
\begin{aligned}
C^{NS}\left(\bm{s}_i,\bm{s}_j;\boldsymbol{\theta}\right)&=\tau\left(\bm{s}_i\right)\tau\left(\bm{s}_j\right)\mathbbm{1}_{{ij}}\left(\bm{s}_i,\bm{s}_j\right) + \frac{\sigma\left(\bm{s}_i\right)\sigma\left(\bm{s}_j\right) 
\left\vert\Sigma\left(\bm{s}_i\right)\right\vert^{1/4}\left\vert\Sigma\left(\bm{s}_j\right)\right\vert^{1/4}}{\Gamma(\bar{\nu}(\bm{s}_i,\bm{s}_j))2^{\bar{\nu}(\bm{s}_i,\bm{s}_j)-1}} \\ 
&\times\left\vert\frac{\Sigma\left(\bm{s}_i\right)+\Sigma\left(\bm{s}_j\right)}{2}\right\vert^{-1/2}  
\left(2\sqrt{{\bar{\nu}(\bm{s}_i,\bm{s}_j)}Q_{ij}}\right)^{\bar{\nu}(\bm{s}_i,\bm{s}_j)}{\cal K}_{\bar{\nu}(\bm{s}_i,\bm{s}_j)}\left(2\sqrt{{\bar{\nu}(\bm{s}_i,\bm{s}_j)}Q_{ij}}\right).
\end{aligned}
\label{eq2}
\end{equation}
Here, $\bar{\nu}(\bm{s}_i,\bm{s}_j) = \frac{\nu(\bm{s}_i)+\nu(\bm{s}_j)}{2} $, $\sigma(\cdot):\mathbb{R}^d\rightarrow\mathbb{R}_+$ is the spatially varying standard deviation, $\tau(\cdot):\mathbb{R}^d\rightarrow\mathbb{R}_+$ is the spatially varying nugget, $\nu(\cdot):\mathbb{R}^d\rightarrow\mathbb{R}_+$ is the spatially varying smoothness, $\Sigma(\cdot):\mathbb{R}^d\rightarrow\mathbb{R}^{d\times d}$ is the spatially varying anisotropy, and $Q_{ij}$ is the Mahalanobis distance between $\bm{s}_i$ and $\bm{s}_j$ such that $Q_{ij}=(\bm{s}_i - \bm{s}_j)^\top \left(\frac{\Sigma(\bm{s}_i)+\Sigma(\bm{s}_j)}{2}\right)^{-1} (\bm{s}_i - \bm{s}_j)$, and ${\cal K}_\nu(\cdot):\mathbb{R}_+\rightarrow\mathbb{R}$ is the modified Bessel function of the second kind with order $\nu>0$. Specifically, for $d=2$, the spatially varying anisotropy can be decomposed as: 
$$
\begin{aligned}
\Sigma(\bm{s}_i) = 
\begin{bmatrix}
\cos(\phi(\bm{s}_i)) & -\sin(\phi(\bm{s}_i)) \\
\sin(\phi(\bm{s}_i)) & \cos(\phi(\bm{s}_i))
\end{bmatrix}
\begin{bmatrix}
\lambda_{1}(\bm{s}_i) & 0 \\
0 & \lambda_{2}(\bm{s}_i)
\end{bmatrix} 
\begin{bmatrix}
\cos(\phi(\bm{s}_i)) & -\sin(\phi(\bm{s}_i)) \\
\sin(\phi(\bm{s}_i)) & \cos(\phi(\bm{s}_i))
\end{bmatrix},
\end{aligned}
$$
where $\phi(\cdot):\mathbb{R}^d\rightarrow (0,\pi/2]$ represents the angle of rotation and $\lambda_{1}(\cdot):\mathbb{R}^d\rightarrow\mathbb{R}_+$,  $\lambda_{2}(\cdot):\mathbb{R}^d\rightarrow\mathbb{R}_+$ are the eigenvalues that represent the spatially varying correlation range. Note that all the covariance parameters in equation~\eqref{eq2} are indeed a function of spatial locations, and therefore, the parameter vector $\boldsymbol{\theta}$ can be redefined as $\boldsymbol{\theta}(\bm{s}_i) =\{ \sigma(\bm{s}_i),\lambda_{1}(\bm{s}_i),\lambda_{2}(\bm{s}_i),\phi(\bm{s}_i),\tau(\bm{s}_i),\nu(\bm{s}_i)\}^\top$. Theoretically, all these parameters can be allowed to vary spatially, which would lead to many rich classes of nonstationarity. However, in favor of the optimization feasibility and detectability issue, certain reasonable restrictions on the spatially varying parameters are considered in practice \citep{anderes2008estimating}.

\subsection{Maximum Likelihood for Nonstationary Mat\'ern Covariance}\label{sec:kernel_smoothing}

Maximum likelihood estimation (MLE) is a commonly employed statistical method for estimating model parameters. Its primary objective is to determine the parameter values in a probability distribution that provide the best fit to the observed data. This is accomplished by maximizing the likelihood function, which quantifies the degree to which the observed data can be explained by the selected distribution and its corresponding parameters. In the context of Gaussian processes (GPs), MLE entails maximizing the Gaussian likelihood function over the parameter space for a given dataset. Specifically, let $\bm{Z} = \{Z(\bm{s}_1),\ldots,Z(\bm{s}_n)\}^\top$ be the observed zero-mean GRF at regularly or irregularly spaced locations $\bm{s}_1,\ldots,\bm{s}_n \in D$. Then $\bm{Z} \sim {\cal N}_n(\bm 0,\bm \Sigma^{NS})$, wherein, $\boldsymbol{\Sigma}^{NS}=\{C^{NS}\left(\bm{s}_i,\bm{s}_j;\boldsymbol{\theta}\left(\bm{s}_i\right),\boldsymbol{\theta}\left(\bm{s}_j\right)\right)\}_{i,j=1}^n$ is the $n\times n$ nonstationary positive-definite covariance matrix. The likelihood can be given as:
\begin{equation}
L\left(\boldsymbol{\theta}\mid\bm{Z}\right)=\frac{1}{\left(2\pi\right)^{n/2}\left|\boldsymbol{\Sigma}^{NS}\right|^{1/2}} \text{exp} \Bigl\{ -\frac{1}{2}\bm{Z}^\top\left(\boldsymbol{\Sigma}^{NS}\right)^{-1}\bm{Z} \Bigl\}.
\label{eq4}
\end{equation}

For parameter estimation, the likelihood function is maximized for the set of parameter vectors, which in case of the nonstationary covariance function $C^{NS}$ is the parameter set $\boldsymbol{\theta}$. Note that we have assumed that the mean function is constant or zero. However, if a more general mean function is considered, a restricted maximum likelihood method may be necessary \citep{patterson1975maximum}. The MLE method for large datasets can be challenging due to the computationally intensive nature of performing large matrix operations. Additionally, when dealing with spatially varying parameters, the task becomes even more difficult due to the presence of a large number of unknown parameters that need to be estimated.

Numerous studies have attempted to tackle this challenge through quicker likelihood approximation methods~\citep{sun2012geostatistics}, such as spectral approximations and composite likelihoods, which makes the optimization computationally feasible on traditional machines. Other approaches rely on advanced parallel linear algebra libraries to support high-performance computing of the MLE operations on a large-scale~\citep{abdulah2018ExaGeoStat}. To tackle the issue of a large number of unknown parameters, we utilize a kernel smoothing approach to represent the spatially varying parameters through smooth functions. This representation has been extensively used in spatial statistics to estimate spatially varying parameters, including the covariance function in Gaussian random fields~\citep{higdon1998process,paciorek2003nonstationary,stein2005nonstationary,li2019efficient}. 
It uses a discrete mixture of the parameters at representative locations called anchor locations through kernel convolution. Specifically, the spatially varying parameters at any given location $\bm{s}_i$ is:
\begin{equation} \label{eq6}
\theta\left(\bm{s}_i\right)=\sum_{k=1}^{K}{W\left(\bm{s}_i,\bm{S}_k\right)\theta\left(\bm{S}_k\right)},
\end{equation}
where $\{\bm{S}_1, \ldots, \bm{S}_K\}$ are the representative locations at $K$ such subregions and 
$W\left(\bm{s}_i,\bm{S}_k\right)=\frac{\mathcal{K}^*\left(\bm{s}_i,\bm{S}_k\right)}{\sum_{k=1}^K\ \mathcal{K}^*\left(\bm{s}_i,\bm{S}_k\right)}$, and the kernel $\mathcal{K}^* \left(\bm{s}_i,\bm{S}_k\right)=\exp\left(-\|\bm{s}_i-\bm{S}_k\|^2/(2h)\right)$ is a Gaussian kernel with bandwidth $h$. The anchor locations can simply be chosen to be the midpoints of each of the sub-regions. The above representation of the spatially varying parameters can capture various spatially varying functions for the covariance parameters. However, the accuracy of the functional form of the spatially varying parameters heavily depends on the choice of the subregions. In this paper, we propose a ConvNet-based subregion selection algorithm to provide better parameter estimates than the existing subjective choices. 

\subsection{Maximum Likelihood Parameter Estimation with ExaGeoStat}

\pkg{ExaGeoStat} \citep{b16,abdulah2018ExaGeoStat,abdulah2019ExaGeoStatr} is a spatial computing software primarily written in C with an R-wrapper available for R-program implementation. It has distributed memory support enabling one to perform parallel computing. Because of this, huge datasets can be incorporated and analyzed through this software. Shared-memory systems, GPUs, and also distributed memory systems can be used through \pkg{ExaGeoStat}. Along with maximum likelihood estimation, simulation of stationary GPs can also be done with \pkg{ExaGeoStat}. We have implemented the nonstationary kernel in \pkg{ExaGeoStat} for large-scale exact simulation of nonstationary random fields along with its parameter estimation, which was previously unavailable in the literature. For our implementation, we have chosen three spatially varying parameters $\{\sigma(\bm{s}_i),\lambda(\bm{s}_i),\nu(\bm{s}_i)\}$ with $\lambda_1 = \lambda_2 = \lambda$ and $\phi$ to be constant.

\section{ConvNet for Nonstationarity Classification} \label{Sec:ConvNet}

Machine learning (ML) has recently garnered significant interest due to its ability to integrate into various applications seamlessly. ML outperforms traditional methods that may encounter challenges in terms of computation or accuracy. It can effectively detect patterns and extract valuable features from data, enabling accurate predictions of missing or future events. Neural network (NN) models are the core of machine learning. These models are designed to mimic the functioning of the human brain, with interconnected layers of artificial neurons that process and analyze data. NN comprises three main layers:  an input layer, one or more hidden layers, and an output layer. Each layer consists of nodes, also known as neurons, associated with specific weights and thresholds. The weights determine the magnitude and impact of the signal transmitted from one node to the next, while the thresholds govern whether a node will activate and forward its output to the subsequent node.

Convolutional neural networks (ConvNet) are a specific type of NN that can effectively process and analyze data with a grid-like structure, such as images or time series data~\citep{rawat2017deep,zhiqiang2017review,ajmal2018convolutional}. Unlike traditional NNs that process data only in a fully connected sequential layers, ConvNets additionally employ convolutional layers that apply filters or kernels to small, overlapping input regions. These filters detect many image features, such as edges, textures, or shapes, by convolving across the input data. This convolutional operation allows the network to capture local dependencies and preserve spatial relationships within the data. This study utilizes ConvNet to classify stationary and nonstationary random fields for spatial data.

\subsection{The ConvNet Architecture}\label{Sec:ConvNet_model}
\begin{figure}[htb]
	\centering
	\includegraphics[width=0.9\textwidth]{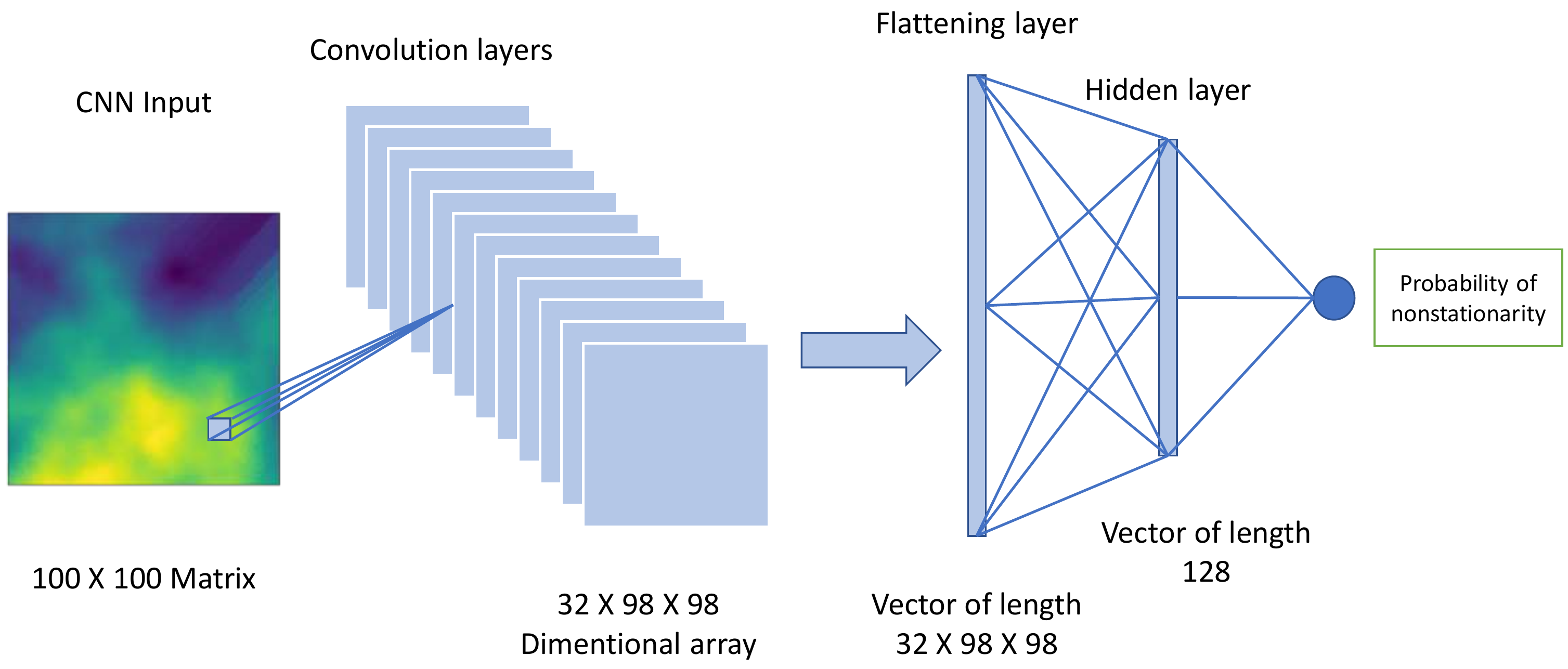}
	\caption{Visualization of the architecture of the proposed ConvNet model. }
	\label{Fig:ConvNet_structure}
\end{figure}

Our proposed ConvNet model effectively distinguishes between stationary and nonstationary regions by taking irregular spatial locations as input. The model generates a probability score indicating the likelihood of a region being stationary. Regions are classified as nonstationary if the probability equals or exceeds 0.5. Conversely, if the probability is below 0.5, the region is categorized as stationary. Figure \ref{Fig:ConvNet_structure} visually illustrates the components of our ConvNet model, which includes the following components:
\begin{itemize}
\item \textbf{Convolution layer}: In this layer, the preprocessed data are fed into a ConvNet through the input layer of dimension $100\times 100$, which captures the spatial neighborhood structure. The input data are derived from a regular grid of size $100\times 100$, and the convolutional layer utilizes a $3\times 3$ kernel. As a result, this layer produces $32$ feature maps, each consisting of $98\times 98$ values in its output. More specifically, let $[\tilde{Z}(i, j)]_{i, j=1}^{100}$ represent the input to this layer. The output of the layer is denoted as $[m_{i, j}^{(\eta)}]_{i, j =1}^{98}$, where $\eta \in \{1, \ldots, 32\}$ corresponds to each feature map. The value of $m_{i, j}^{(\eta)}$ is determined by the equation:
	\begin{equation*}
		m_{i, j}^{(\eta)} = \sigma\left( \sum_{r=1}^3\sum_{s=1}^3  \tilde{Z}(i+r-1, j +s-1) k_{r,s}^{(\eta)} + b^{(\eta)} \right), 
	\end{equation*}
where, $\sigma(x) = \max(0, x)$ represents the Rectified Linear Unit (RELU) activation function. The kernel matrix $[k_{r, s}^{(\eta)}]_{r,s=1}^3\in\mathbb{R}^{3\times 3}$ and the bias $b^{(\eta)}\in\mathbb{R}$, $\eta=1,\ldots,32,$ are specific to this layer, resulting in a total of $(3\times 3 + 1) \times 32 = 320$ parameters.

\item \textbf{Flatten layer}: The role of this layer is to convert the outputs from the preceding convolutional layer into a flattened vector of length $32\times 98\times 98$. Specifically, the tensor of dimension $32\times 98\times 98$ is concatenated into a single vector of length $N_F=32\times 98 \times 98=307,328$.

\item \textbf{Fully connected layer}: In this layer, the flattened vector from the previous layer undergoes two sequential nonlinear transformations. Firstly, the incoming vector of length $N_F$ is transformed into a hidden layer vector of length $H_1=128$ through an elementwise RELU activation operation on the affine transformation with weights and biases. Similarly, it is further transformed into a vector of length $H_2=2$. The vector input to Hidden Layer 1 is denoted as $[\tilde{m}_i]_{i=1}^{N_F}$, and the output is represented by $[\tilde{m}_i^{(1)}]_{i=1}^{H_1}$. The computation of $[\tilde{m}_i^{(1)}]_{i=1}^{H_1}$ is given as follows:
	\begin{equation*}
		\tilde{m}_i^{(1)} = \sigma\left( \sum_{j=1}^{N_F} p_{i, j}^{(1)} \tilde{m}_j + p_{0, j}^{(1)} \right), 
	\end{equation*}
where $[p_{i, j}^{(1)}]_{1\leq i \leq H_1, 1\leq j \leq N_F}\in\mathbb{R}^{H_1\times N_F}$ and $[p_{0, i}^{(1)}]_{i=1}^{H_1}$ represent the parameters in terms of weights and biases, respectively. The total number of parameters in this layer is $(N_F + 1) \times H_1$.

Moving up to Hidden Layer 2, the input is $[\tilde{m}_i^{(1)}]_{i=1}^{H_1}$, and the output is $[\tilde{m}_i^{(2)}]_{i=1}^{H_2}$, where the computation of the outputs is given by:
\begin{equation*}
\tilde{m}_i^{(2)} = \sigma\left( \sum_{j=1}^{H_1} p_{i, j}^{(2)} \tilde{m}_j^{(1)} + p_{0, j}^{(2)} \right),
\end{equation*}

where $[p_{i, j}^{(2)}]_{1\leq i \leq H_2, 1\leq j \leq H_1}\in\mathbb{R}^{H_2\times H_1}$ and $[p_{0, i}^{(1)}]_{i=1}^{H_2}$ represent the parameters in terms of weights and biases, respectively. The total number of parameters in this layer is $(H_1 + 1) \times H_2$.
\item \textbf{Output layer}: The purpose of this layer is to generate the nonstationarity index by transforming the vector of length $2$. The inputs to this layer are $\tilde{m}_1^{(2)}$ and $\tilde{m}_2^{(2)}$, and the output is obtained through the following expression:
	\begin{equation*}
		I_{nonstat} = \frac{\exp(\tilde{m}_1^{(2)})}{\exp(\tilde{m}_1^{(2)}) + \exp(\tilde{m}_2^{(2)})}, 
	\end{equation*}
	which is the probability of the random field being nonstationary. 
\end{itemize}

\subsection{The ConvNet Training Phase and Data Pre-processing}\label{Sec:Data_Preprocessing}

In order to train the ConvNet model for classifying stationary and nonstationary data, we leverage the \texttt{TensorFlow} library. During the training process, we utilize the categorical cross-entropy function as the loss function and the Adam optimizer. The model undergoes training for 25 epochs. After completion of the training, the ConvNet model provides a nonstationarity probability for each spatial dataset. We have developed the ConvNet model as a standalone approach, allowing it to be employed for classifying any stationary or nonstationary spatial field. Further details and implementation of Python code can be found at \url{https://github.com/kaust-es/Spatial_classifier_DL.git}.
This development can serve as a standalone tool for classifying stationary and nonstationary random fields.

\begin{figure}[t!]
	\centering
	\includegraphics[width=\textwidth]{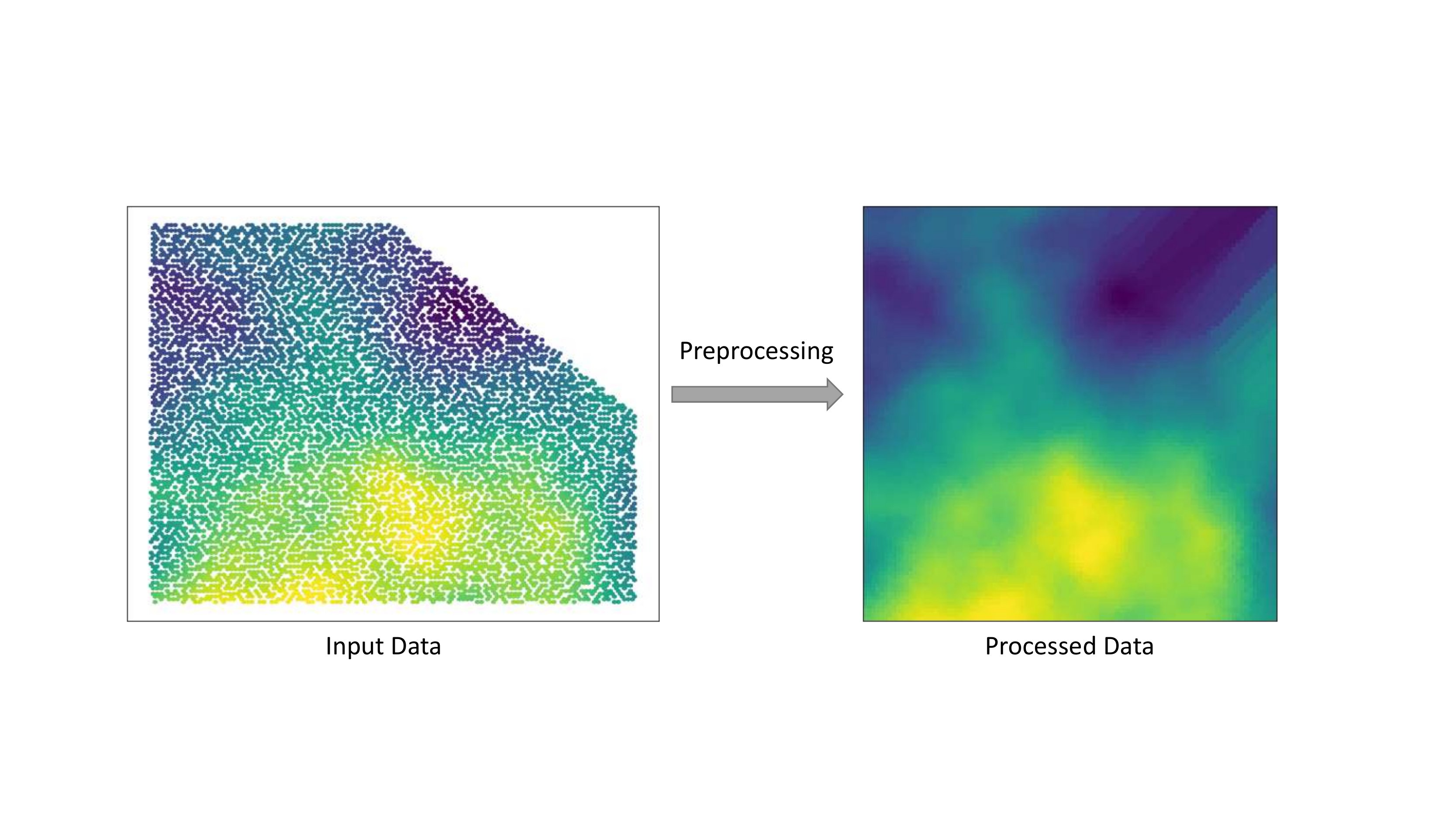}
	\caption{Data pre-processing step.}
	\label{Fig:Preprocessing_step}
\end{figure}

Training the ConvNet discussed in Section~\ref{Sec:ConvNet_model} can be directly accomplished with regularly gridded datasets. However, real-world applications often involve irregularly spaced spatial data. To address this, we have developed a data preprocessing stage. Figure \ref{Fig:Preprocessing_step} illustrates the idea behind the pre-processing step.


To pre-process the spatial field $Z(\bm{s})$ observed at $n$ locations, recall the raw data as $\bm{Z} = (Z(\bm{s}_1), \ldots, Z(\bm{s}_n))^\top$, where $\bm{s}_i \in D$. Our goal is to map the locations from $D$ to $[0, 1]^2$, assuming that the observed region can be properly stretched. The resulting pre-processed data are $\tilde{\bm{Z}} = [\tilde{Z}(i, j)]_{i, j= 1}^{100}$, which represent spatial data observed on a $100\times 100$ regular grid. The steps can be broken down into the following:
\begin{itemize}
	\item \textbf{Splitting:} We divide the observed region $[0, 1]^2$ to $100\times 100$ subregions indexed by $(i, j)$ equally. Here the subregions $D_{i, j}$ are defined by
	\begin{equation*}
		D_{i, j} = \begin{cases}
			[(i-1)/100, i/100) \times [(j-1)/100, j/100), & i, j < 100; \\
			[99/100, 1] \times [(j-1)/100, j/100), & i=100, j < 100; \\
			[(i-1)/100, i/100) \times [99/100, 1], & i < 100, j = 100; \\
			[99/100, 1] \times [99/100, 1], & i, j = 100. 
		\end{cases}
		\label{eq:preprocess}
	\end{equation*}
	\item \textbf{Averaging:} Let $N_{i, j}$ be the total number of observations within $\bm{s}_1, \ldots, \bm{s}_n$ that lie within the subregion $D_{i, j}$. We define $\bar{Z}(i, j)$ as the mean of the obervations within $D_{i, j}$. More specifically, we define 
	\begin{equation*}
		\bar{Z}(i, j) = \begin{cases}
			\frac{1}{N_{i, j}} \sum_{k: \bm{s}_k \in D_{i, j}} Z(\bm{s}_k), & N_{i, j} > 0; \\
			\bar{Z}(u, v), \text{ where } (u, v) = \operatorname*{argmin}_{u,v} \{(i-u)^2 + (j -v)^2 : N_{u, v} > 0 \} & N_{i, j} = 0. 
		\end{cases}
	\end{equation*}
	\item \textbf{Scaling:} The averaging results $\bar{Z}(i, j)$ are scaled such that the range of the observations is $[0, 1]$. Let $m = \min_{i, j \in {1, \ldots, 100}} \bar{Z}(i, j)$, $M = \max_{i, j \in {1, \ldots, 100}} \bar{Z}(i, j)$. We have 
	\begin{equation*}
		\tilde{Z}(i, j) = \begin{cases}
			(\bar{Z}(i, j) - m) / (M - m), & m \neq M; \\
			0.5, & m = M. 
		\end{cases}
	\end{equation*}
\end{itemize}

\subsection{ConvNet for Subregion Selection} \label{Sec:clustering}

In this section, we introduce a data-driven subregion partition method via ConvNet. The proposed algorithm requires a predetermined number of subregions, denoted by $K$. For a randomly selected $K$ points, $\mathbf{A} = \{\bm{a}_1,\ldots,\bm{a}_K\}$ from $\mathbf{S}_{vec} = \{\bm{s}_1,\ldots,\bm{s}_n\}$, we obtain the vector of distances 
$
\mathbf{E} = \{\|\bm{s}_i-\bm{a}_k\|^2 : \bm{a}_k \in \mathbf{A}\}, \bm{s}_i \in \mathbf{S}_{vec}. 
$ We then assign $(\bm{s}_i,Z(\bm{s}_i))$ to subregion $\mathcal{R}_k$, if $\mathbf{E}_k$ is minimum in $\mathbf{E}$. 
Next, we preprocess the data for each subregion and use ConvNet to determine the probability of being nonstationary. The sum of such probabilities for all subregions, denoted by $P=\sum_{k=1}^Kp_k$, indicates the degree of nonstationarity. We repeat this whole procedure $iters$ number of times and select the set of subregions with the lowest $P$ as the final partition for estimating the nonstationary covariance function. The anchor locations $\mathbf{S}$ are then defined to be the centre points of those subregions. Algorithm~\ref{alg:alg1} summarizes the whole procedure.


\begin{algorithm}

\caption{ConvNet Subregion Selection Algorithm }
\label{alg:alg1}
\begin{algorithmic}
\Require A set of locations $\mathbf{S}_{vec}$, and observations $\mathbf{Z} = \{Z(\bfs_1),\ldots, Z(\bfs_n)\}$,  number of subregions $K$, and number of iterations $iters$

\Ensure Anchor locations $\mathbf{S} = \{\mathbf{S}_1,\ldots,\mathbf{S}_K\}$ for $K$ subregions

\For{\texttt{$u = 1 : iters$}}
    \State Select $K$ random points $\mathbf{A}$ from the given set of locations $\mathbf{S}_{vec}$ 
    \For{\texttt{$i = 1 : n$}}
        \State Calculate $\mathbf{E}$ for $\bm{s}_i$
        \State Assign $(\bm{s}_i,Z(\bm{s}_i))$ to $\mathcal{R}_k^u$ if $\mathbf{E}_k$ is minimum 
    \EndFor    
    \Comment{$K$ subregions $\{\mathcal{R}_1^u,\ldots,\mathcal{R}_K^u\}$ for $u$-th iteration}
    
    \For{\texttt{$k = 1 : K$}}
        \State $\tilde{\bm{Z}}_k = PreProcess(\mathcal{R}_k^u)$
        \Comment{Follow \eqref{eq:preprocess}}
        \State $p_k = \text{ConvNet}(\tilde{\bm{Z}}_k)$
        \Comment{Nonstationary probability of each region $k$}
    \EndFor
    
    $P = \sum_{k=1}^K p_k$
    \If{$min > P $}
        \State $min = P$  and $index = u$ 
        \State $\{\mathcal{R}_1,\ldots,\mathcal{R}_K\}=\{\mathcal{R}_1^u,\ldots,\mathcal{R}_K^u\}$
    \EndIf

\EndFor
\State $\mathbf{S} = \{(m(\mathcal{R}_1),\ldots,m(\mathcal{R}_K)\}$
\Comment{$m(\mathcal{R}_k)$ is the centre of the subregion $\mathcal{R}_k$}
\end{algorithmic}
\end{algorithm}


The algorithm has been implemented within the \pkg{ExaGeoStat} software using C++. To import the ConvNet model trained in \texttt{Tensorflow} into \pkg{ExaGeoStat}, we utilized the \pkg{Tensorflow\_C\_API}, which is an open-source software for integrating Python models into C/C++. Additionally, we seamlessly integrated the C++ code into the C environment of \pkg{ExaGeoStat} using the \texttt{CppFlow} package. During the installation of the package, the ConvNet model is automatically downloaded and placed in the appropriate folder to enable the execution of the algorithm. By leveraging this algorithm, \pkg{ExaGeoStat} can obtain subregions and provide parameter estimates and predictions for the nonstationary kernel.

\section{Simulation Studies} \label{Sec:Experiment}
This section presents a set of experiments to achieve two primary objectives. Firstly, we evaluate the effectiveness of our proposed ConvNet model in classifying a given spatial region as either stationary or nonstationary, providing an associated probability value. Secondly, we assess the effectiveness of the proposed ConvNet-based nonstationary parameter estimation method and compare its quality to fixed-partitioning methods. We rely on the \pkg{ExaGeoStat} framework to conduct large-scale modeling experiments. 


\subsection{Settings of the Synthetic Spatial Datasets}

To validate the effectiveness of our proposed ConvNet model in classifying spatial data regions, we utilized the spatial data generation tool within the \pkg{ExaGeoStat} framework to generate a set of synthetic training and testing datasets. The locations within the datasets were generated using the formula $n^{-1/2}(i - 0.5 + X_{i, j}, j - 0.5 + Y_{i, j})$, where $i, j \in {1, \ldots, n^{1/2}}$ and the values of $X_{i, j}$ and $Y_{i, j}$ were generated from independent and identically distributed (i.i.d.) samples of the uniform distribution $\text{Unif}(-0.4, 0.4)$. This process ensured that the spatial locations of the data followed a specific pattern for analysis. Therefore, the observed region is $[0, 1]^2$. All datasets were generated using the isotropic Matérn random field for the stationary case, while for the nonstationary case, they were generated from a spatial process with spatially varying variance. In total, we generated $16{,}000$ stationary datasets and $16{,}000$ nonstationary datasets. From these, we randomly selected 12,800 stationary datasets and $12{,}800$ nonstationary datasets for training the ConvNet model, while the remaining data were reserved for testing purposes. The generated synthetic datasets underwent preprocessing before the training and testing phases, following the method outlined in Section \ref{Sec:Data_Preprocessing}. As a result, the data meets the necessary criteria for our ConvNet model. For ConvNet model training, the number of observations is $n=10{,}000$. Similar to \cite{abdulah2018ExaGeoStat}, the stationary datasets are generated from a stationary isotropic Gaussian random field with Mat\'{e}rn covariance: 
\begin{equation} \label{Eq:Matern_Covariance}
	{\cal M}(h;, \bm{\theta}) = \frac{\sigma^2}{2^{\nu-1} \Gamma(\nu)}\left(\frac{h}{\alpha}\right)^\nu {\cal K}_\nu \left(\frac{h}{\alpha}\right), 
\end{equation}
where $\bm{\theta} = (\sigma^2, \alpha, \nu)^\top>\boldsymbol{0}$. The effective range $h_{\text{eff}}$ is defined as the value such that ${\cal M}(h_{\text{eff}};\bm{\theta}) /\sigma^2 = 0.05$. 
We set $\sigma^2=1$, $\nu = 1/8 + (3/16)k$ for $k \in \{0, 15\}$, and $\alpha$ is chosen such that $h_{\text{eff}} = 0.05 + 0.003k$ for $k \in \{0, \ldots, 999\}$. For each parameter combination, we generate one sample for the data, so there are $16{,}000$ stationary data samples. 

The nonstationary datasets are generated from a random field with spatially varying variance. This process is defined by 
\begin{equation*}
	Z(\bm{s}) = (0.01 + 0.99\sigma^{(u)}(\bm{s})) Z_S(\bm{s}), 
\end{equation*}
where $\sigma^{(u)}(\bm{s})$ is the spatially varying terms with $u \in \{1, \ldots, 5\}$ for different patterns, and $Z_S(\bm{s})$ is a stationary Gaussian random field with Mat\'{e}rn covariance \eqref{Eq:Matern_Covariance}. For each pattern of $\sigma^{(u)}$, there are 3,200 nonstationary data samples, so in total there are 16,000 nonstationary data samples. We generate one sample for each parameter combination. Letting $\bm{s} = (x_1, x_2)$, details of the nonstationary patterns are provided as follows: 
\begin{itemize}
	\item For $u=1$: 
	\begin{equation*}
		\sigma^{(1)}(\bm{s}) = \frac{1}{2} \sin \left[ r_A \{(x_1-0.5)\cos(\theta_A) + (x_2-0.5)\sin(\theta_A)\} \right] + \frac{1}{2}, 
	\end{equation*}
	where $r_A \in \{5, 10, 15, \ldots, 50\}$, $\theta_A \in \{k\pi/12: k = 0, 1, \ldots, 11\} \cup \{\pi/4 + k\pi/2, k = 0, 1, 2, 3\}$. In random field $Z_S(\bm{s})$, we choose $\sigma^2=1$, $\nu \in \{0.5, 1, 1.5, 2\}$, and $h_{\text{eff}} \in \{0.1, 0.2, 0.4, 0.8, 1.6\}$. 
	\item For $u=2$: 
	\begin{equation*}
		\sigma^{(2)}(\bm{s}) = \frac{1}{2} \sin \left[ \left[ r_B \{(x_1-0.5)\cos(\theta_B) + (x_2-0.5)\sin(\theta_B)\} \right]^{2p_B} \right] + \frac{1}{2}, 
	\end{equation*}
	where $p_B \in \{1, 2, 3\}$, $r_B \in \{3, 4, \ldots, 10\}$, or $p_B = 4$, $r_B = 3$; $\theta_B \in \{k\pi/12: k = 0, 1, \ldots, 11\} \cup \{\pi/4 + k\pi/2, k = 0, 1, 2, 3\}$. In random field $Z_S(\bm{s})$, we choose $\sigma^2=1$, $\nu \in \{0.5, 1\}$, and $h_{\text{eff}} \in \{0.2, 0.4, 0.8, 1.6\}$. 
	\item For $u=3$: 
	\begin{equation*}
		\sigma^{(3)}(\bm{s}) = \frac{ \exp\left[ \sin\{r_C (x_1-0.5)\cos(\theta_C) \} + \sin\{r_C (x_2-0.5)\sin(\theta_C) \} \right] - e^{-2}}{e^2-e^{-2}}, 
	\end{equation*}
	where $r_C$, $\theta_C$, and parameter combinations of $Z_S(\bm{s})$ are similar to the case of $u=1$. 
	\item For $u=4$: 
	\begin{equation*}
		\sigma^{(4)}(\bm{s}) = \frac{ (x_1-0.5)\cos(\theta_D) + (x_2-0.5)\sin(\theta_D)}{|\cos(\theta_D)| + |\sin(\theta_D)|} + \frac{1}{2}, 
	\end{equation*}
	where $\theta_D \in \{k\pi/12: k = 0, 1, \ldots, 11\} \cup \{\pi/4 + k\pi/2, k = 0, 1, 2, 3\}$. In random field $Z_S(\bm{s})$, we choose $\sigma^2=1$, $\nu \in \{k/4: k=1, \ldots, 8\}$, and $h_{\text{eff}} \in \{0.1 + k/16: k = 0, \ldots, 24\}$. 
	\item For $u=5$: We first define 
	\begin{equation*}
		\begin{bmatrix}
			x_1^\star \\
			x_2^\star
		\end{bmatrix} = 
		\begin{bmatrix}
			0.5 \\
			0.5
		\end{bmatrix} + 
		\begin{bmatrix}
			\cos \theta_E & - \sin \theta_E \\
			\sin \theta_E & \cos \theta_E
		\end{bmatrix}
		\begin{bmatrix}
			x_1-0.5 \\
			x_2-0.5
		\end{bmatrix} 
	\end{equation*}
	and $\bar{x}^\star = (x_1^\star + x_2^\star) / 2$. Let 
	\begin{equation*}
		\sigma_0^{(5)}(\bm{s}) = 3\sin \{20 (\bar{x}^\star + 1.9)\} \cos \{ 20 (\bar{x}^\star - 1.2)^6\} + 0.6e^{ \sin(25x_1^\star) + \sin (13x_2^\star)} + \frac{\bar{x}^\star - 0.2}{2}, 
	\end{equation*}
	and $\sigma^{(5)}(\bm{s})$ is obtained by scaling $\sigma_0^{(5)}(\bm{s})$ such that the minimum and maximum of $\sigma^{(5)}(\bm{s})$ are $0$ and $1$ for all observation locations, respectively. Here $\theta_E$ and parameter combinations of $Z_S(\bm{s})$ are similar to the case of $u=4$. Along with the spatially varying $\sigma$ we also used the nonstationary kernel as discussed in Section \ref{nonstat_matern} for simulation of nonstationary random fields with spatially varying parameters $\{\sigma(\bm{s}),\lambda(\bm{s}),\nu(\bm{s})\}$. 
\end{itemize}

\subsection{ConvNet Model Accuracy Assessment}

We used $12{,}800$ data samples from both the stationary and nonstationary regime to train our ConNet model. Subsequently, we evaluated the model's accuracy by testing it on the remaining dataset, consisting of $3200$ stationary and $3200$ nonstationary samples generated during the data generation phase.

\begin{figure}[htb]

\centering
\includegraphics[width=\textwidth]{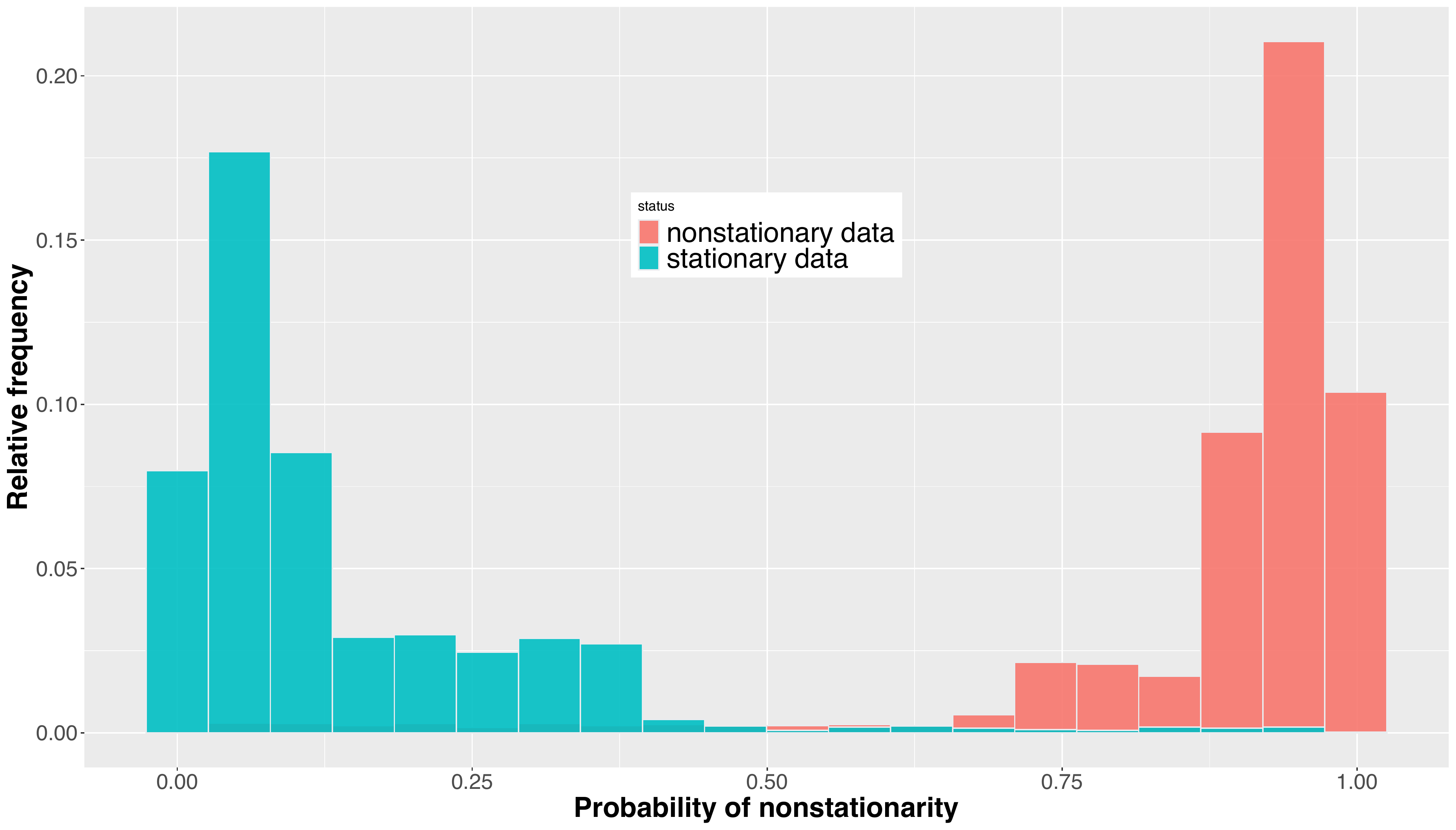}

	\caption{Histogram of the nonstationarity index for stationary testing data and nonstationary testing data generated from spatially varying models.}
	\label{Fig:Hist_Spatially_Varying} 
\end{figure}

Figure \ref{Fig:Hist_Spatially_Varying} illustrates two histograms. The left histogram represents the accuracy of our proposed ConvNet model when applied to stationary data. It demonstrates that the model can successfully detect stationarity (low probability of nonstationarity) in 97.3\% of the provided datasets. The right histogram showcases the model's exceptional ability to accurately identify nonstationary spatial data (high probability of nonstationarity), achieving a high percentage of 98.4\%. Overall, our trained model achieved an impressive accuracy of 97.9\% in correctly classifying the stationary and nonstationary datasets.

\subsection{Parameter Estimation Quality}

Once the ConvNet model is trained, it becomes capable of identifying low-probability nonstationary regions. This information is then used to perform splits of spatial regions to obtain the representative anchor locations from detected subregions. We rely on the \pkg{ExaGeoStat} software to estimate the parameters at all anchor locations. To evaluate the performance of the estimation based on the given subregions, we conducted simulations in three different settings, focusing on estimating three parameters: $\sigma(\bm{s})$, $\lambda(\bm{s})$, and $\nu(\bm{s})$. For each simulation scenario, we present the average values of the parameter estimates across replicated simulations.

\begin{figure}[htp!]
\captionsetup[subfigure]{labelformat=empty}
   \begin{minipage}[t]{\linewidth}
       \vspace{3mm}
   \subcaption{Setting 1 (Data generated with four subregions)}
 \vspace{1mm}
 \begin{subfigure}[b]{\textwidth}
\centering
\includegraphics[width=0.2\textwidth]{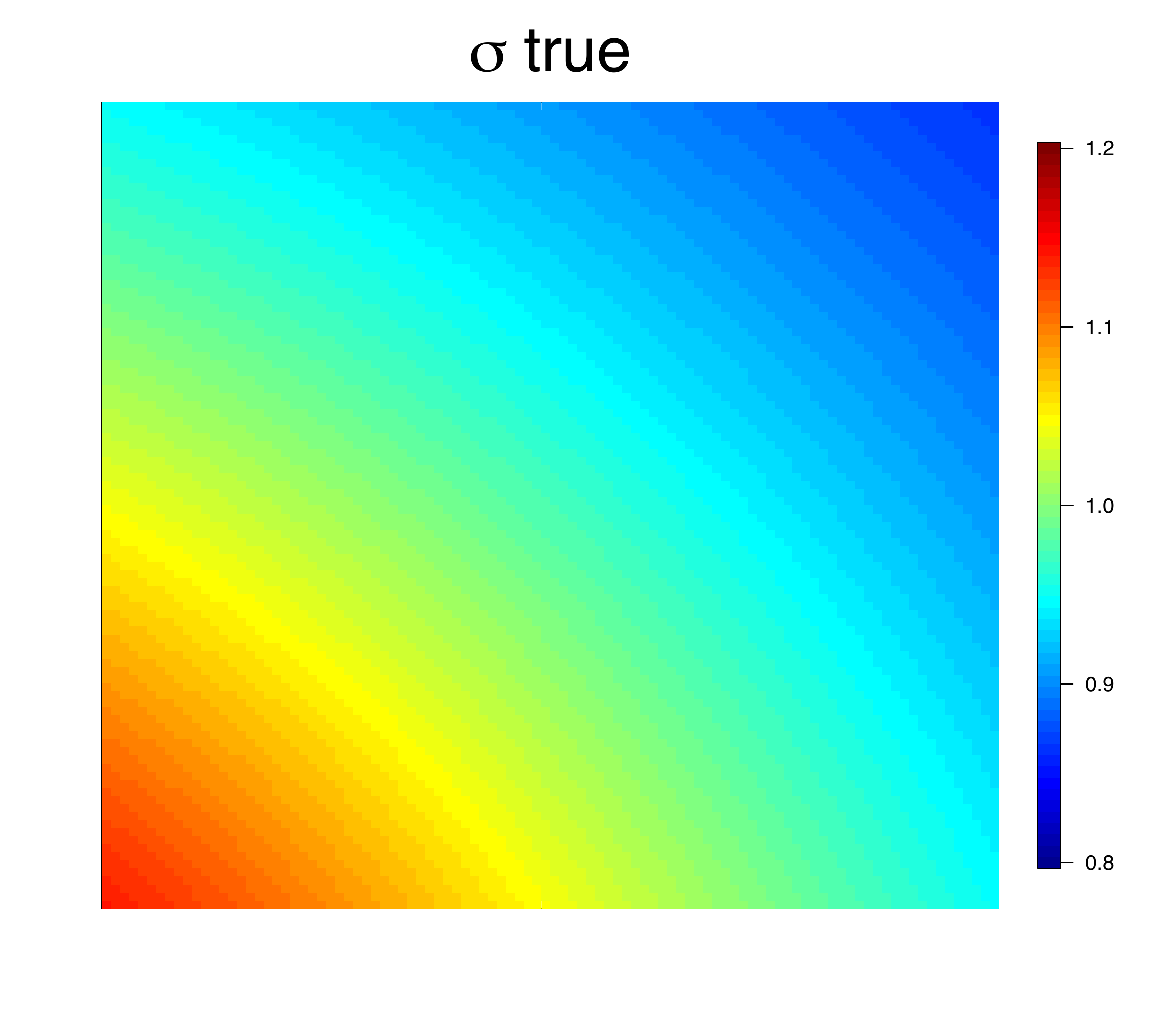}
\hspace{-2mm}
\includegraphics[width=0.2\textwidth]{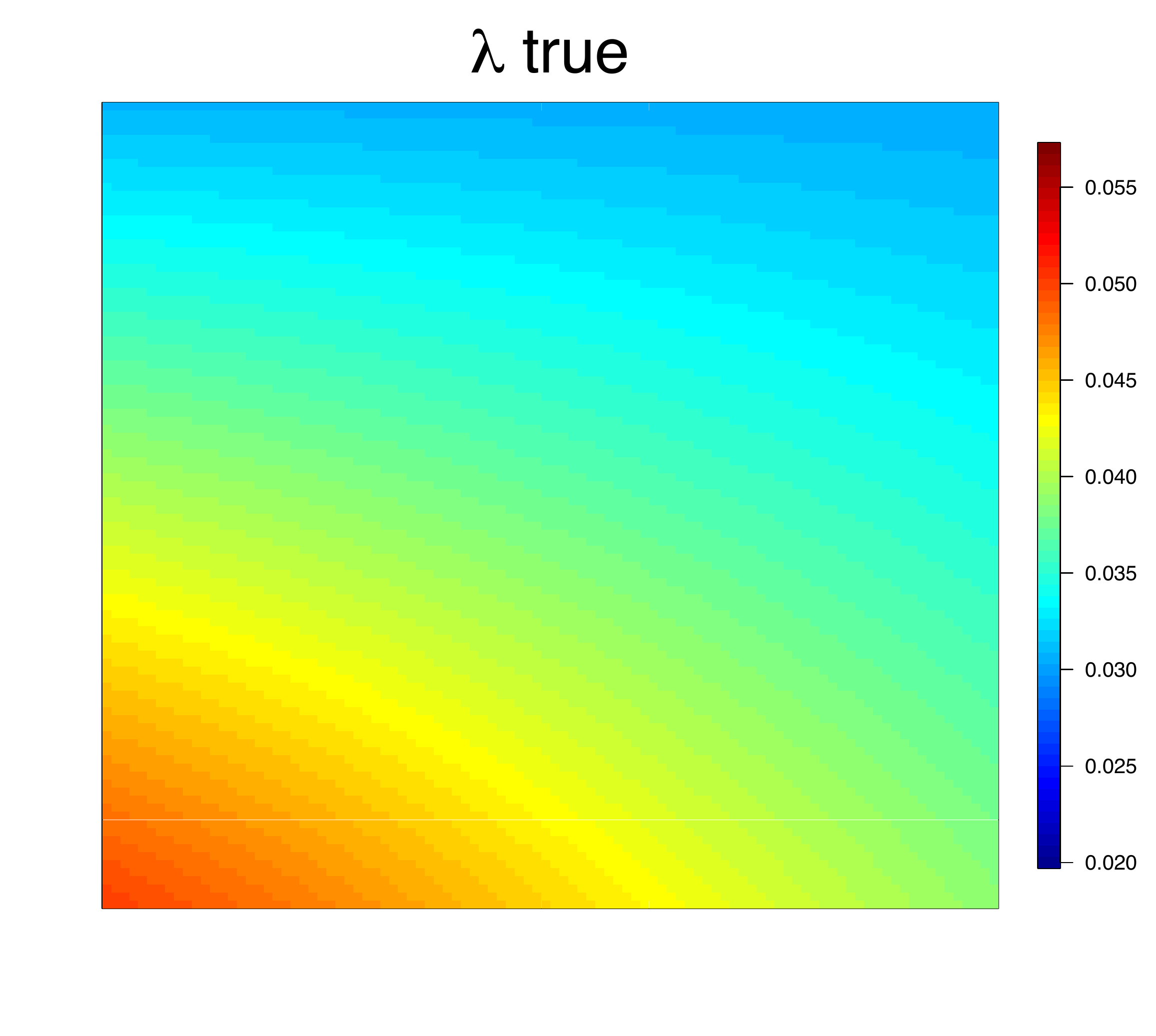}
\hspace{-2mm}
\includegraphics[width=0.2\textwidth]{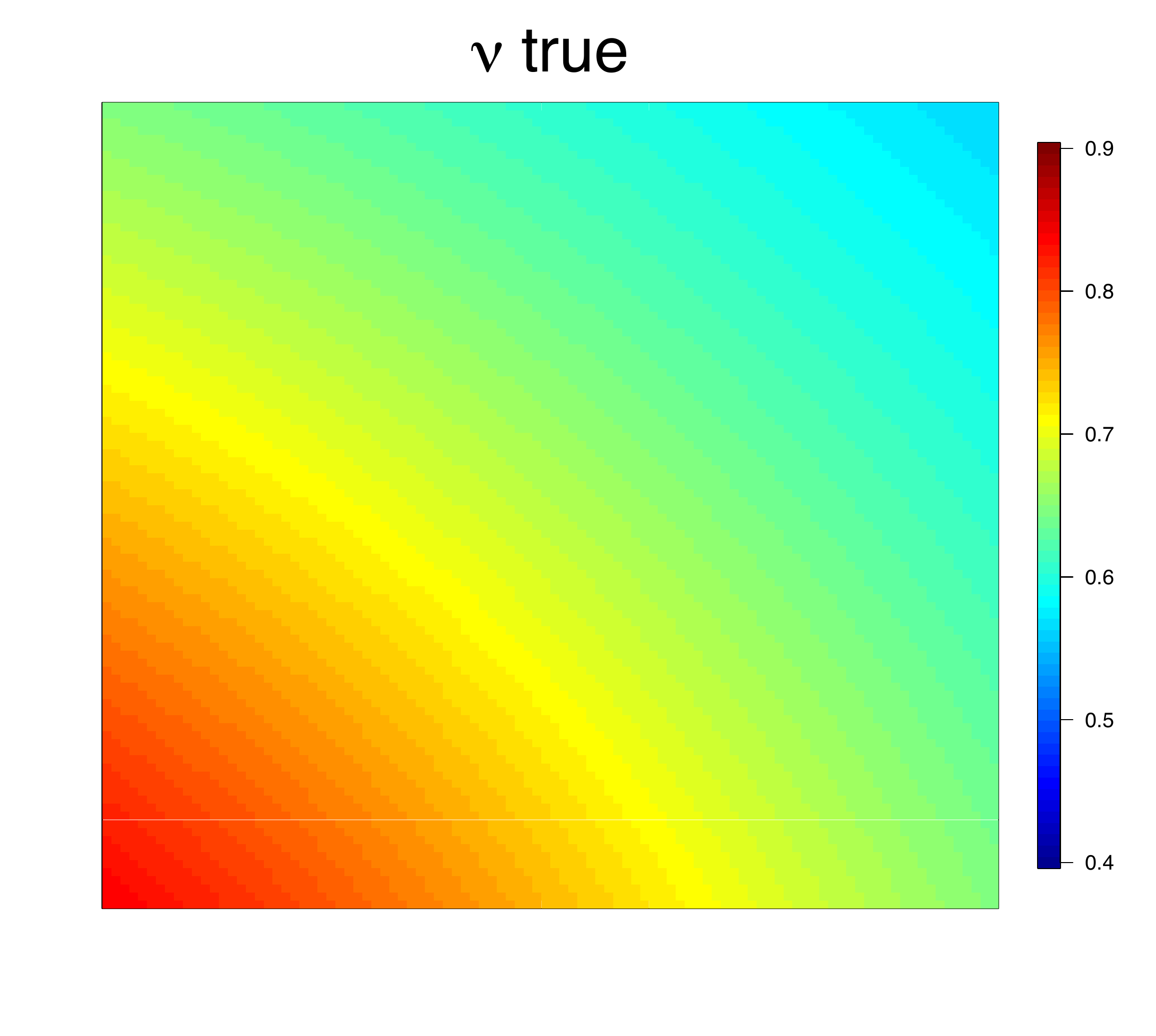}
\vspace{-4mm}
\caption{True parameters} 
\vspace{3mm}
 \end{subfigure}
\end{minipage}
   \begin{minipage}[t]{\linewidth}
       \vspace{3mm}
 \vspace{-2mm}
\centering
\begin{subfigure}[b]{0.46\textwidth}
\centering
\includegraphics[width=0.32\textwidth]{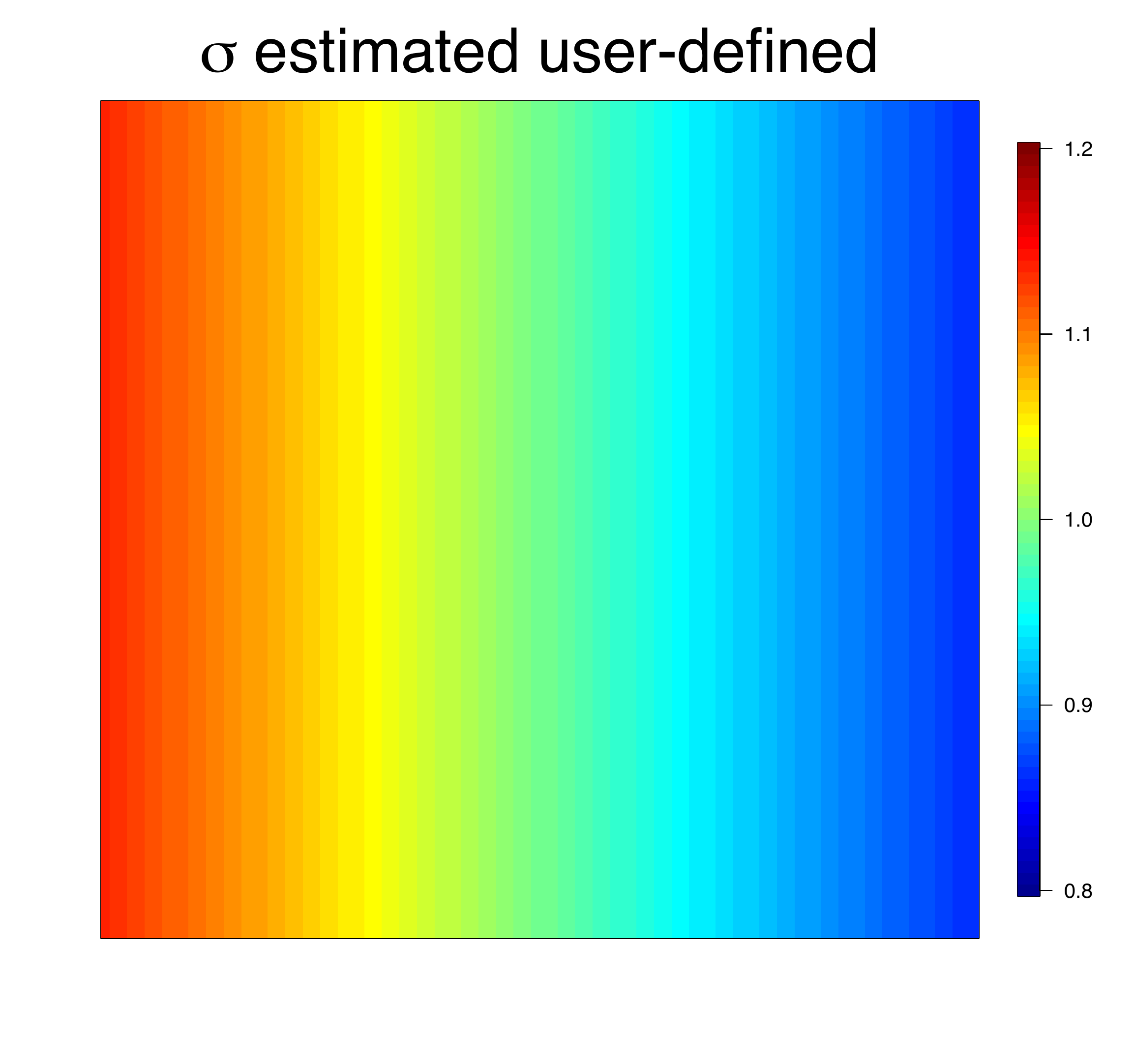}
\hspace{-2mm}
\includegraphics[width=0.32\textwidth]{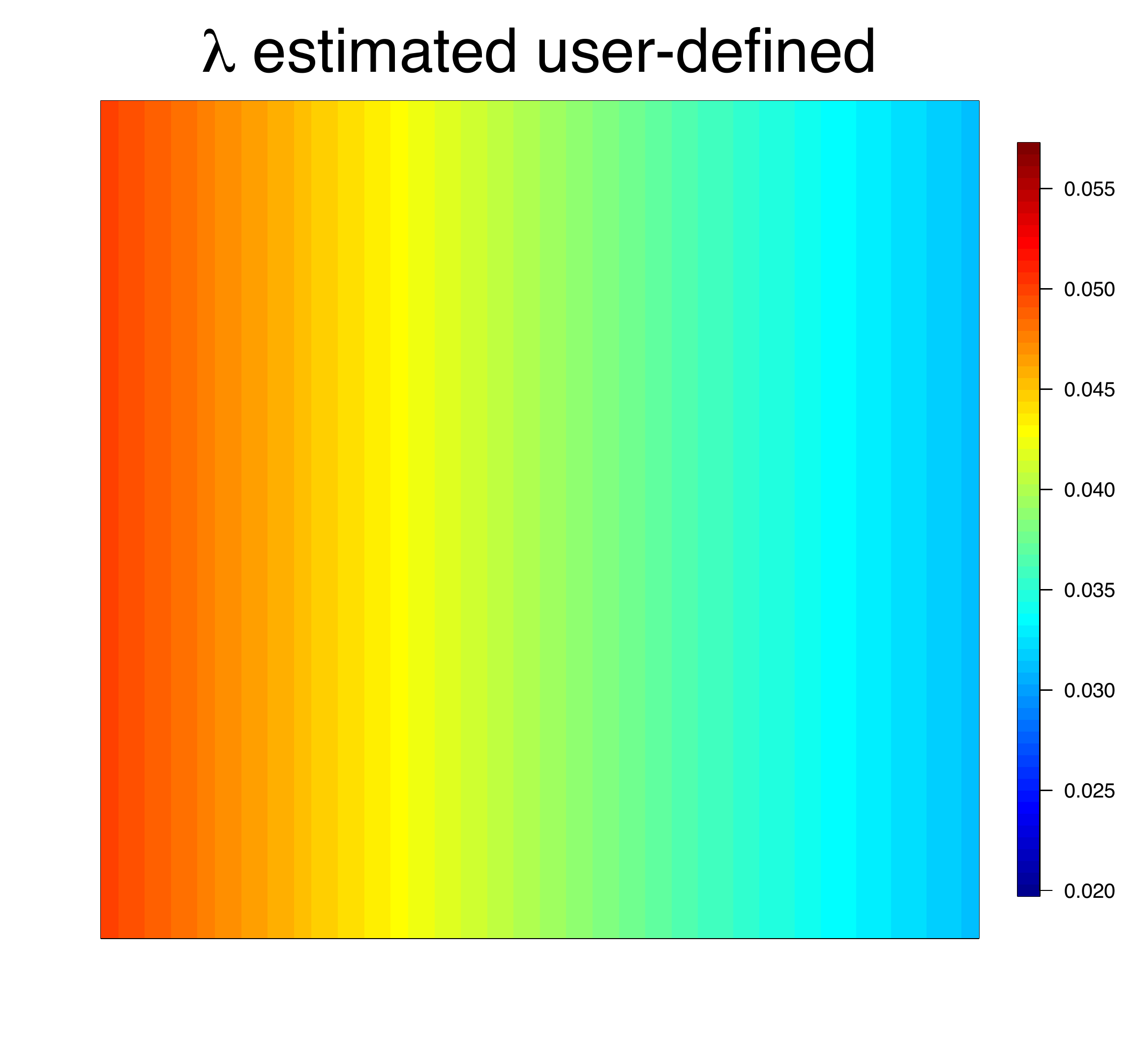}
\hspace{-2mm}
\includegraphics[width=0.32\textwidth]{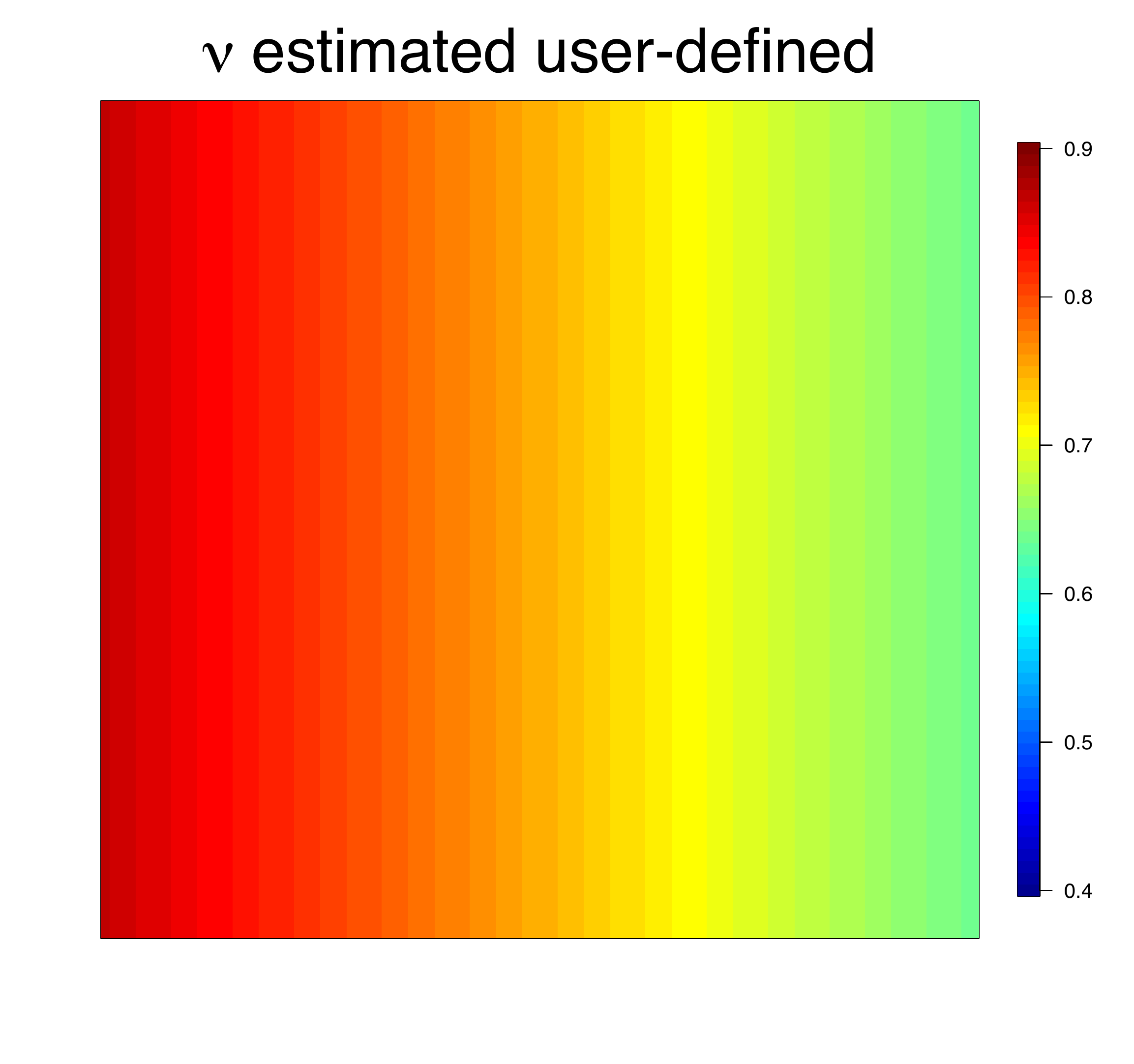}
\vspace{-4mm}
\caption{Estimated with three user-defined subregions. } 
\vspace{4mm}
 \end{subfigure}
  \hspace{8mm}
\begin{subfigure}[b]{0.46\textwidth}
\centering
\includegraphics[width=0.32\textwidth]{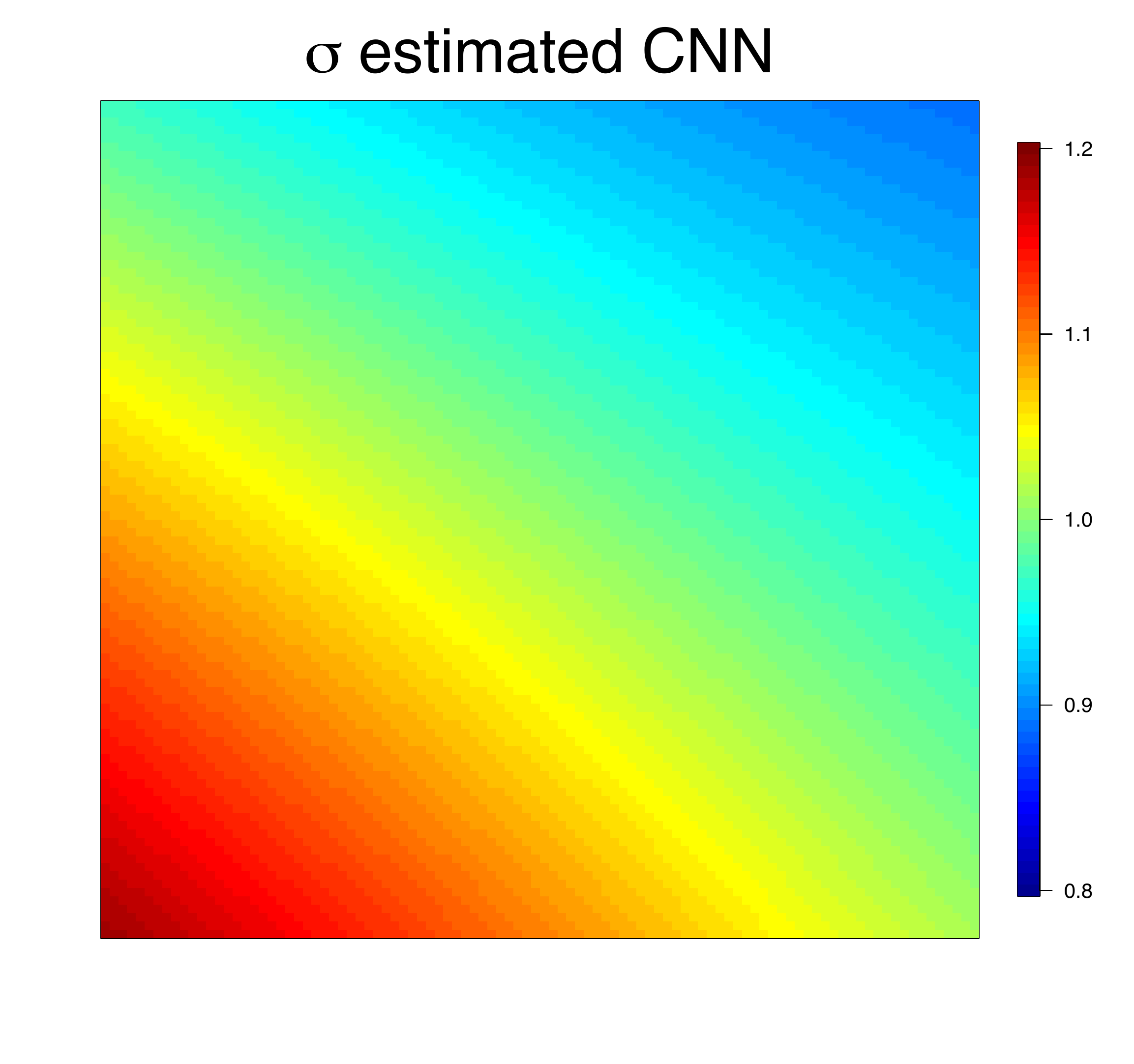}
\hspace{-2mm}
\includegraphics[width=0.32\textwidth]{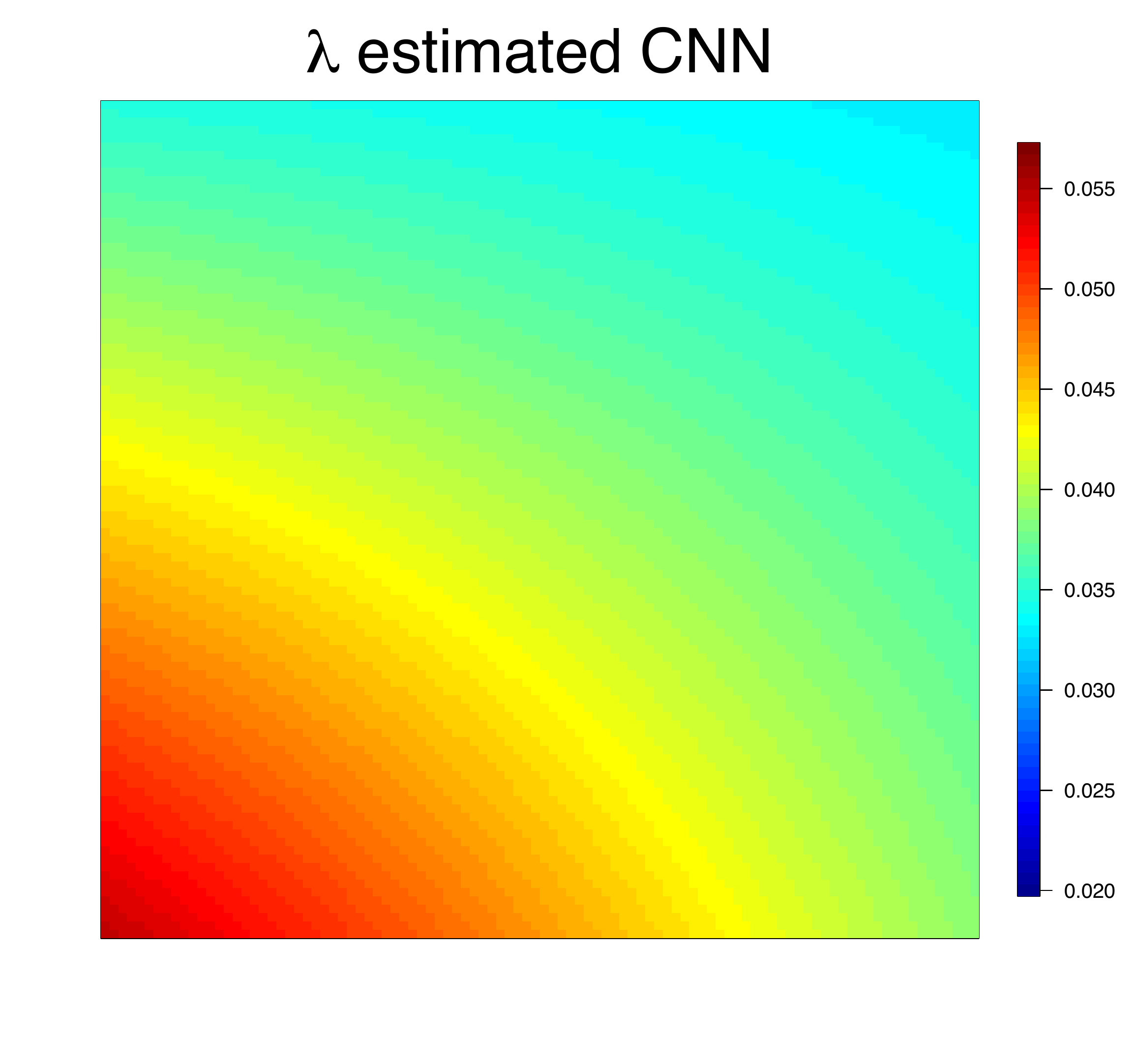}
\hspace{-2mm}
\includegraphics[width=0.32\textwidth]{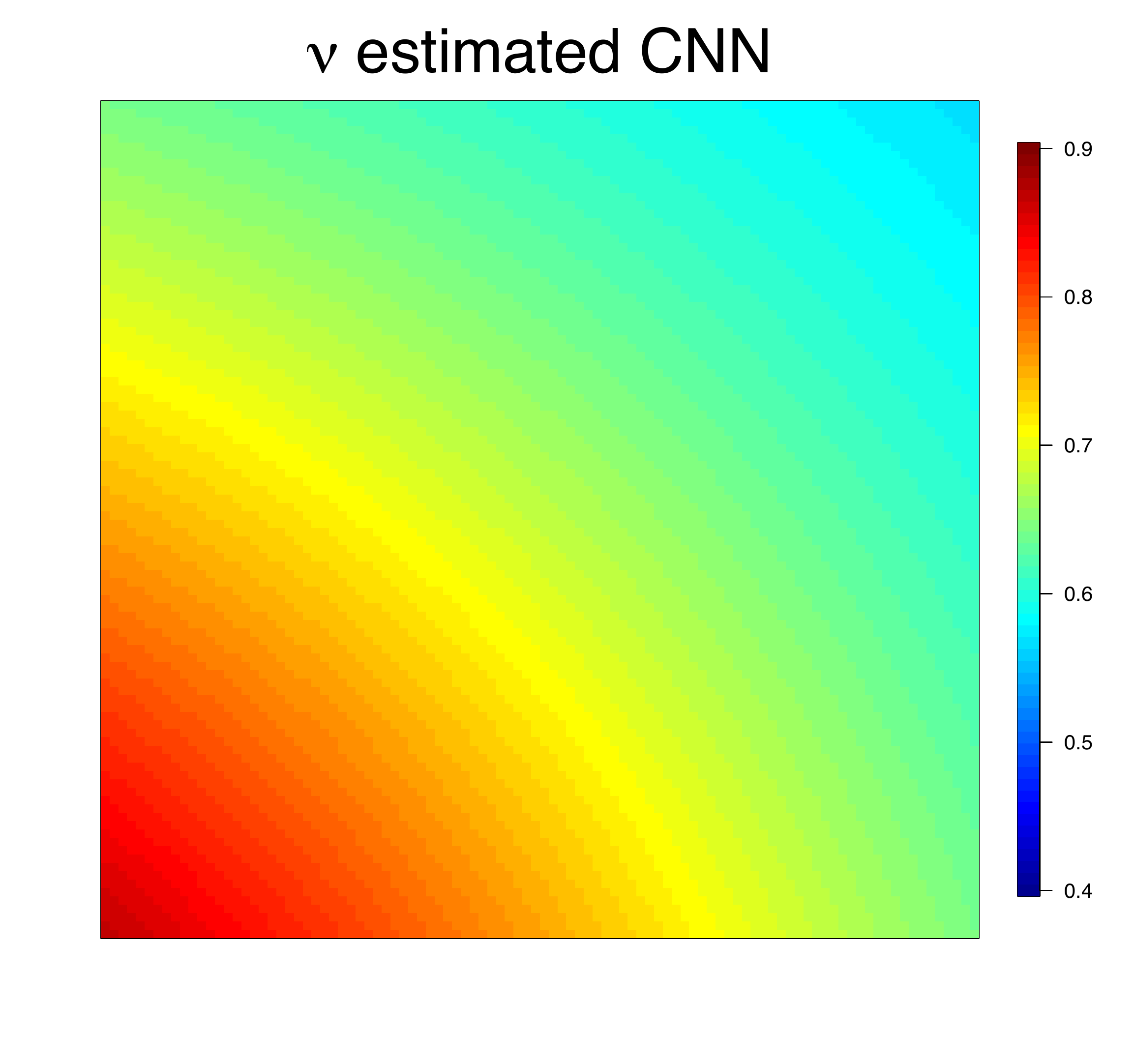}
\vspace{-4mm}
\caption{Estimated with three ConvNet subregions.} 
\vspace{4mm}
 \end{subfigure}
\end{minipage}

   \begin{minipage}[t]{\linewidth}
       \vspace{3mm}

 \vspace{-2mm}
\centering
\begin{subfigure}[b]{0.46\textwidth}
\centering
\includegraphics[width=0.32\textwidth]{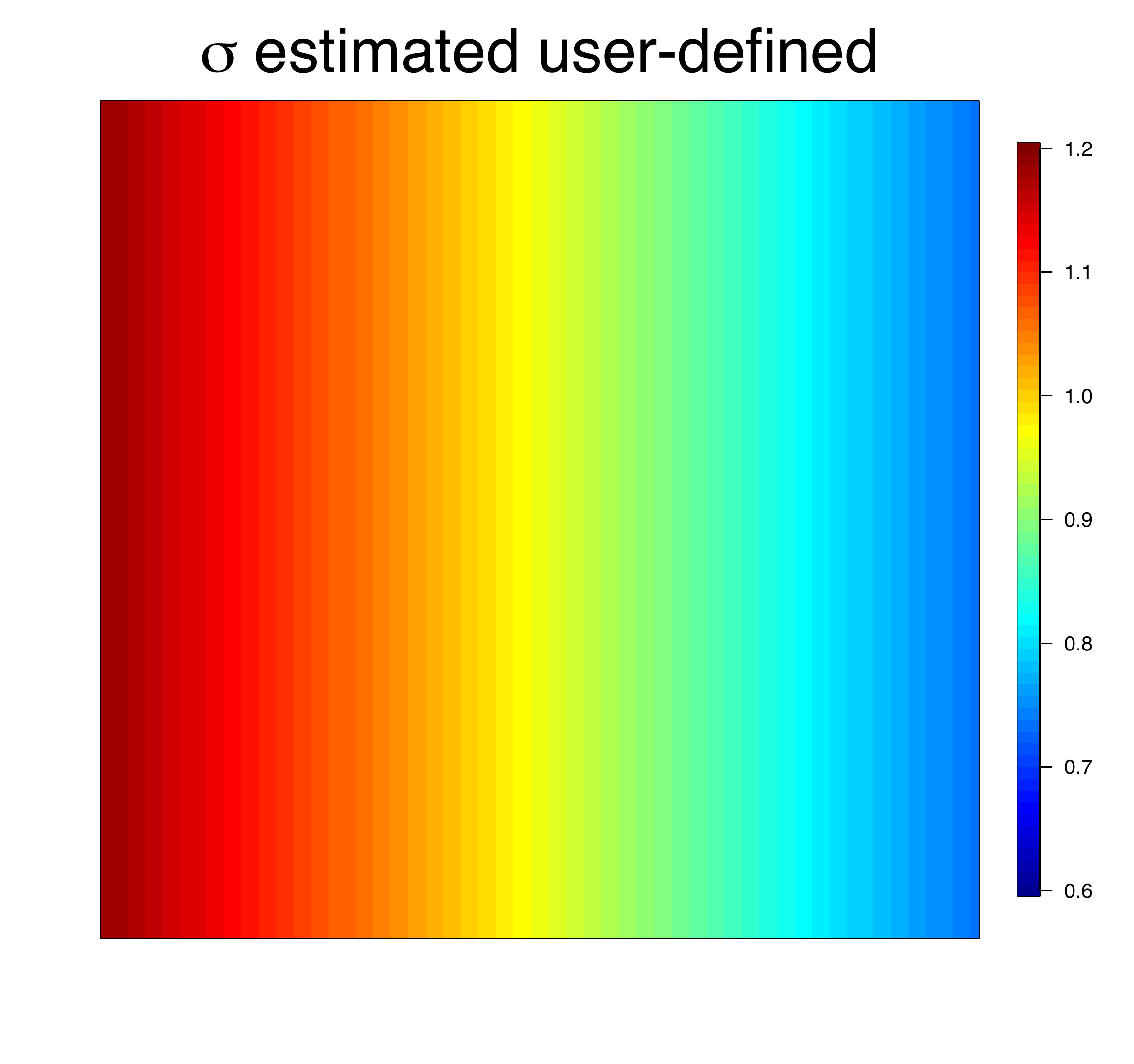}
\hspace{-2mm}
\includegraphics[width=0.32\textwidth]{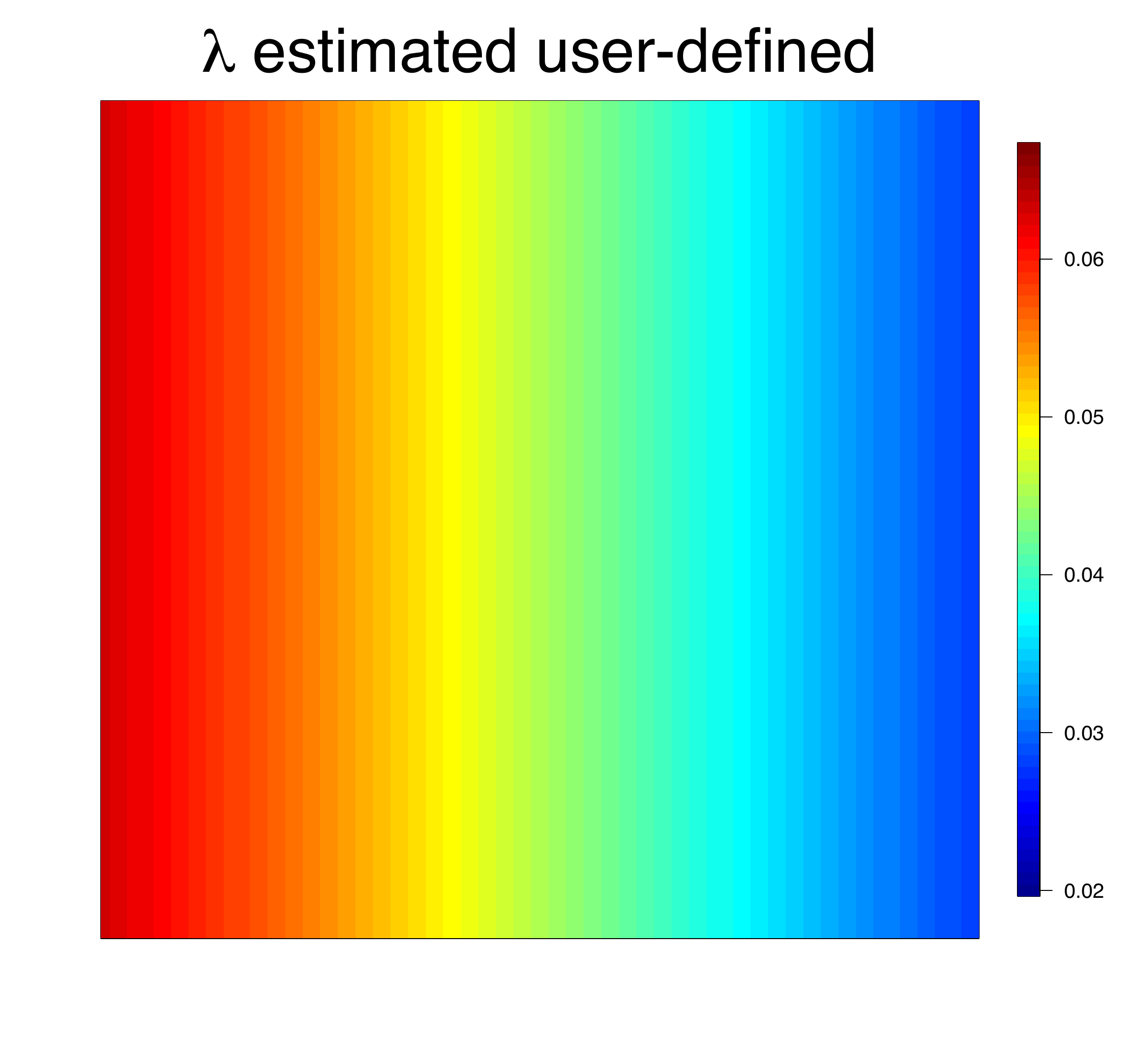}
\hspace{-2mm}
\includegraphics[width=0.32\textwidth]{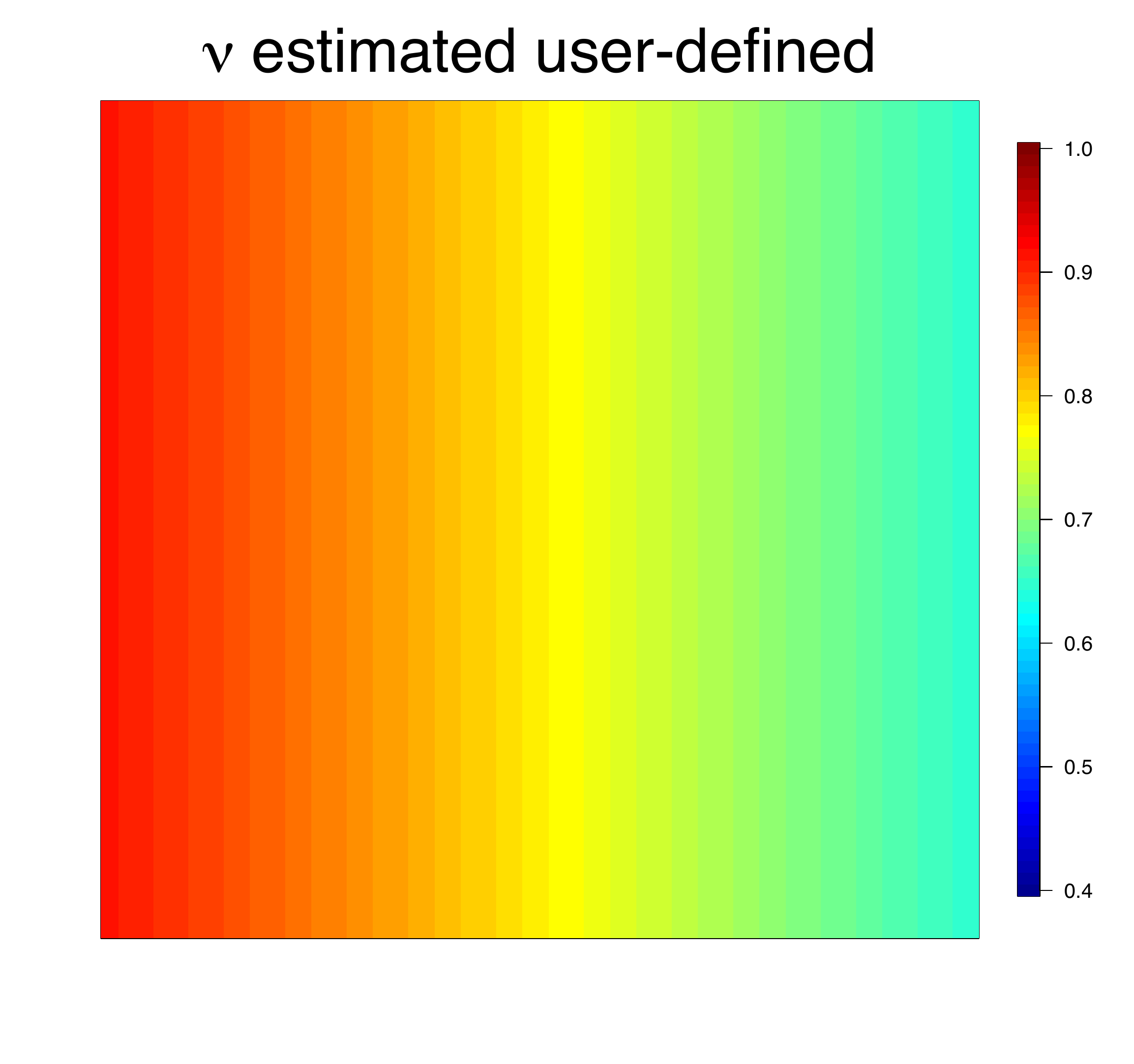}
\vspace{-4mm}
\caption{Estimated with two user-defined subregions. } 
\vspace{4mm}
 \end{subfigure}
  \hspace{8mm}
\begin{subfigure}[b]{0.46\textwidth}
\centering
\includegraphics[width=0.32\textwidth]{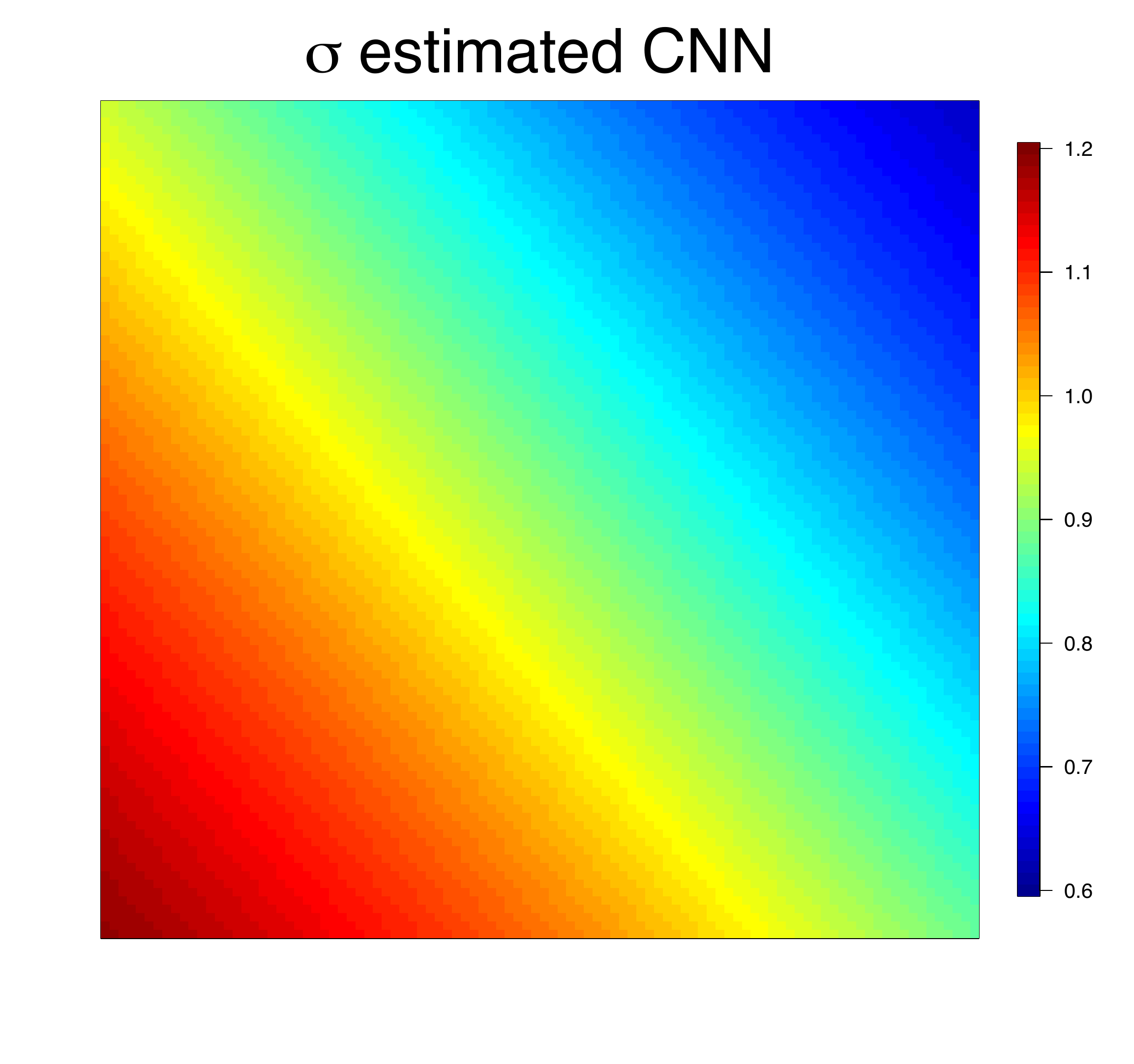}
\hspace{-2mm}
\includegraphics[width=0.32\textwidth]{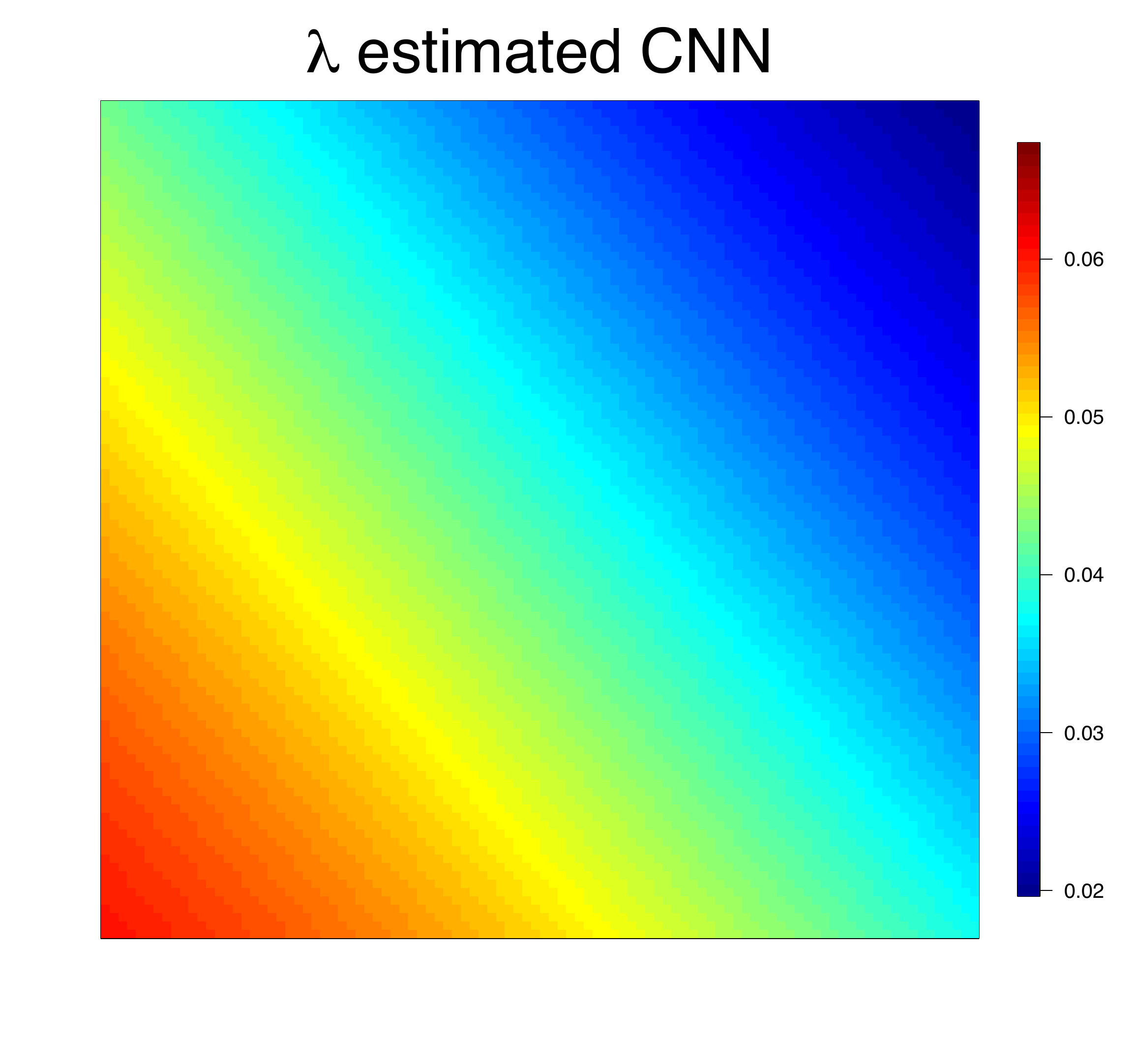}
\hspace{-2mm}
\includegraphics[width=0.32\textwidth]{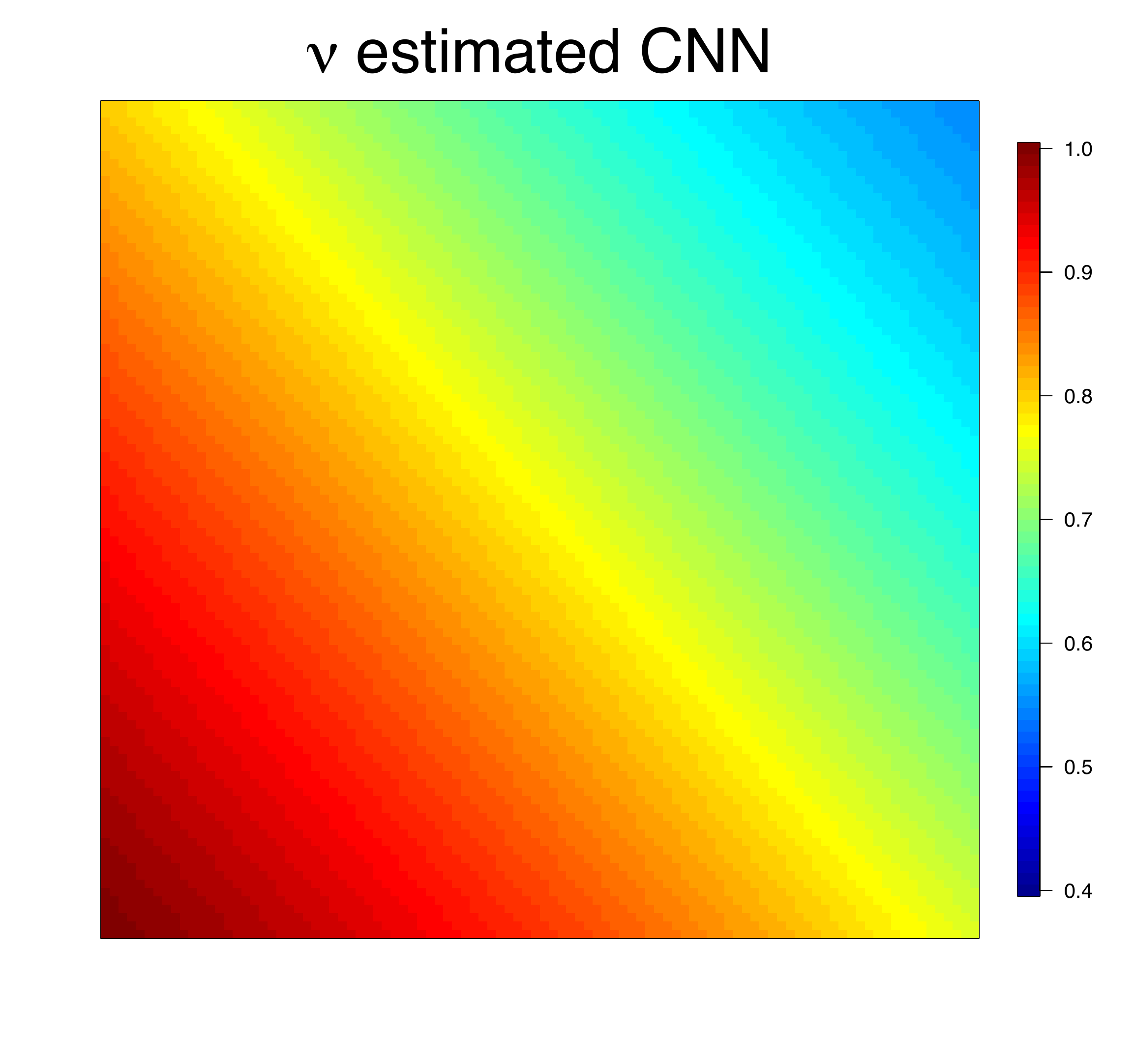}
\vspace{-4mm}
\caption{Estimated with two ConvNet subregions.} 
\vspace{4mm}
 \end{subfigure}
\end{minipage}

\begin{minipage}[t]{\linewidth}
       \vspace{10mm}
   \subcaption{Setting 2 (Data generated with two subregions)}
 \vspace{1mm}
 \begin{subfigure}[b]{\textwidth}
\centering
\includegraphics[width=0.2\textwidth]{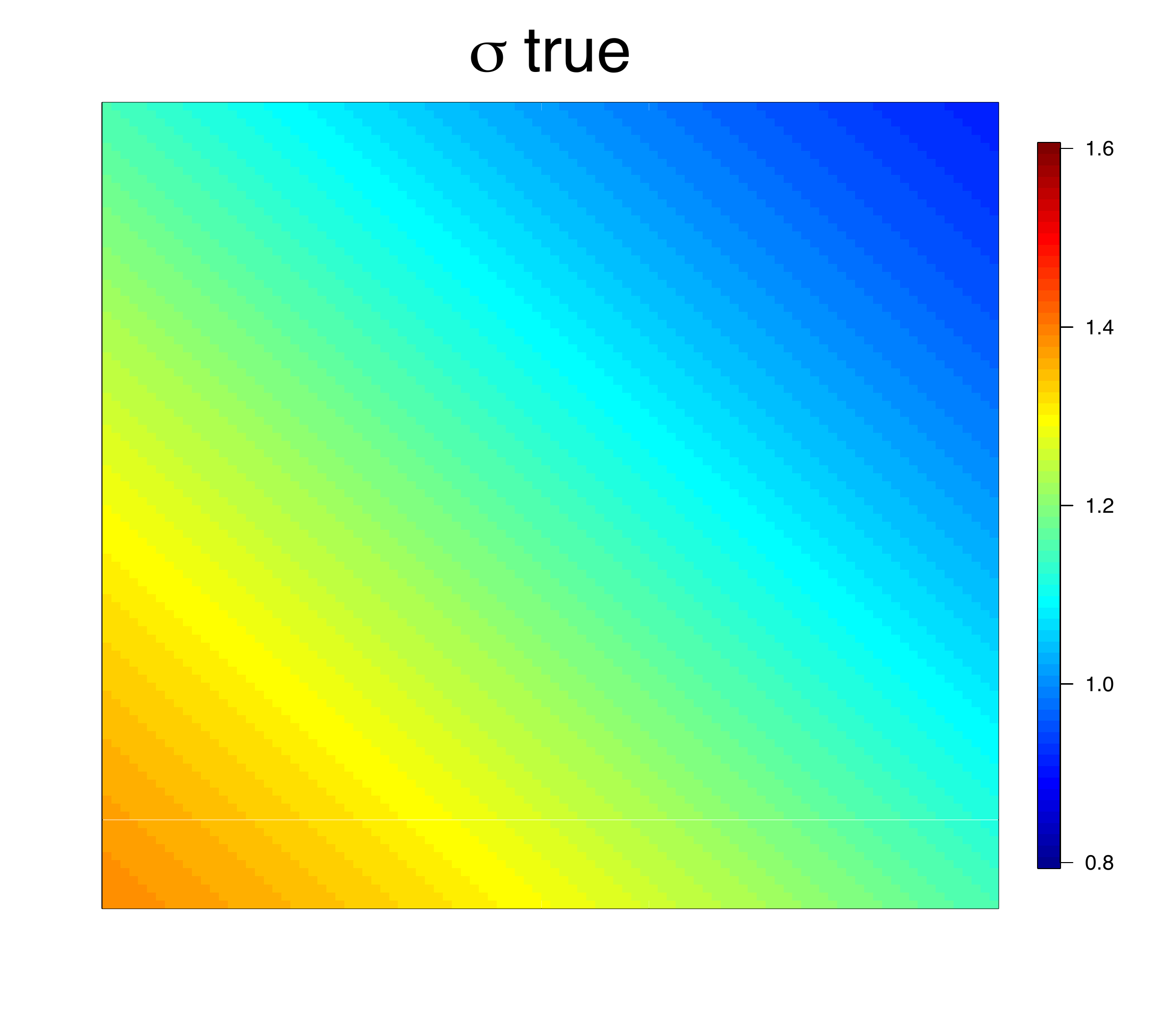}
\hspace{-2mm}
\includegraphics[width=0.2\textwidth]{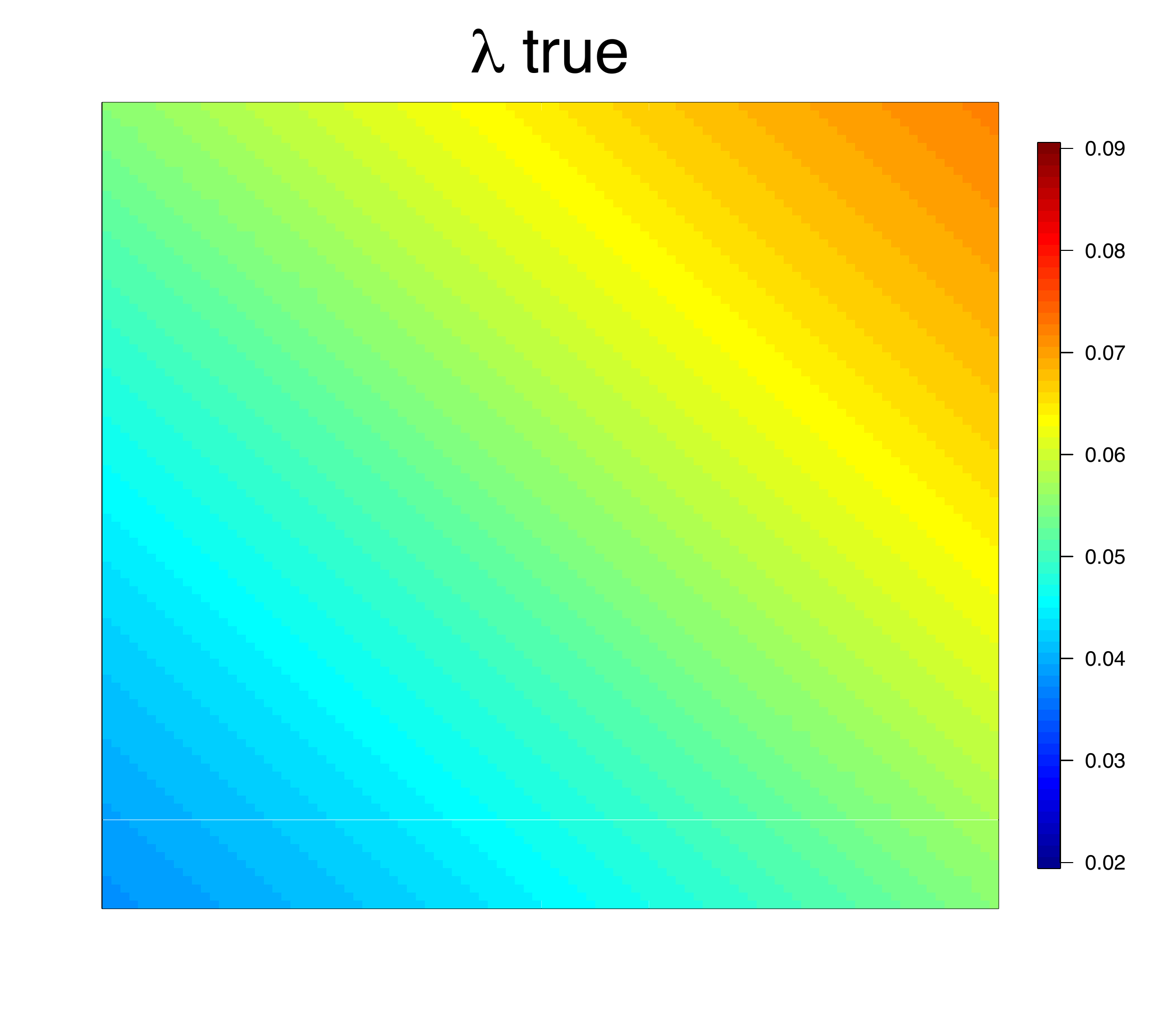}
\hspace{-2mm}
\includegraphics[width=0.2\textwidth]{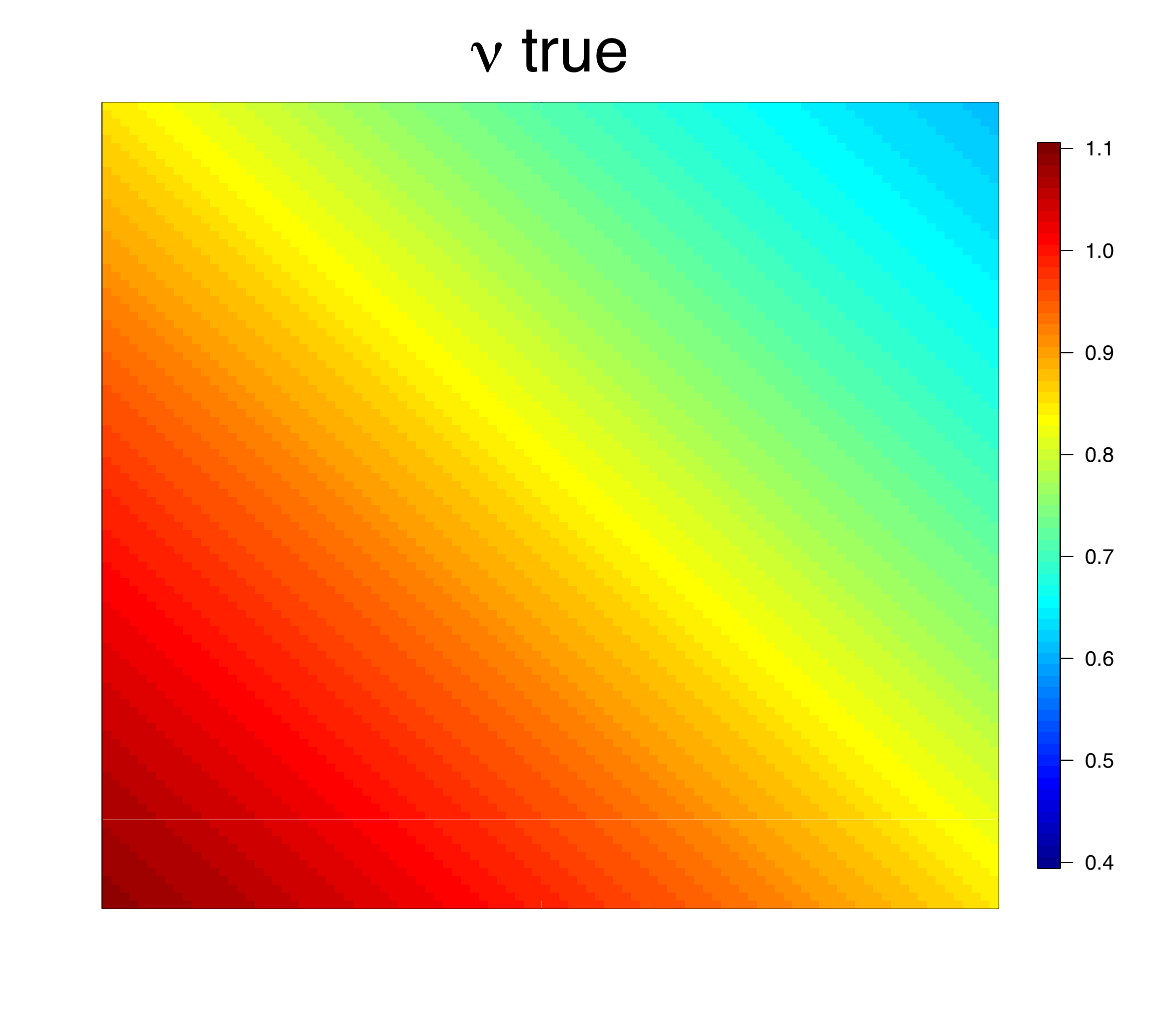}
\vspace{-4mm}
\caption{True parameters} 
\vspace{4mm}
 \end{subfigure}
\end{minipage}

   \begin{minipage}[t]{\linewidth}
       \vspace{3mm}
 \vspace{-2mm}
\centering
\begin{subfigure}[b]{0.46\textwidth}
\centering
\includegraphics[width=0.32\textwidth]{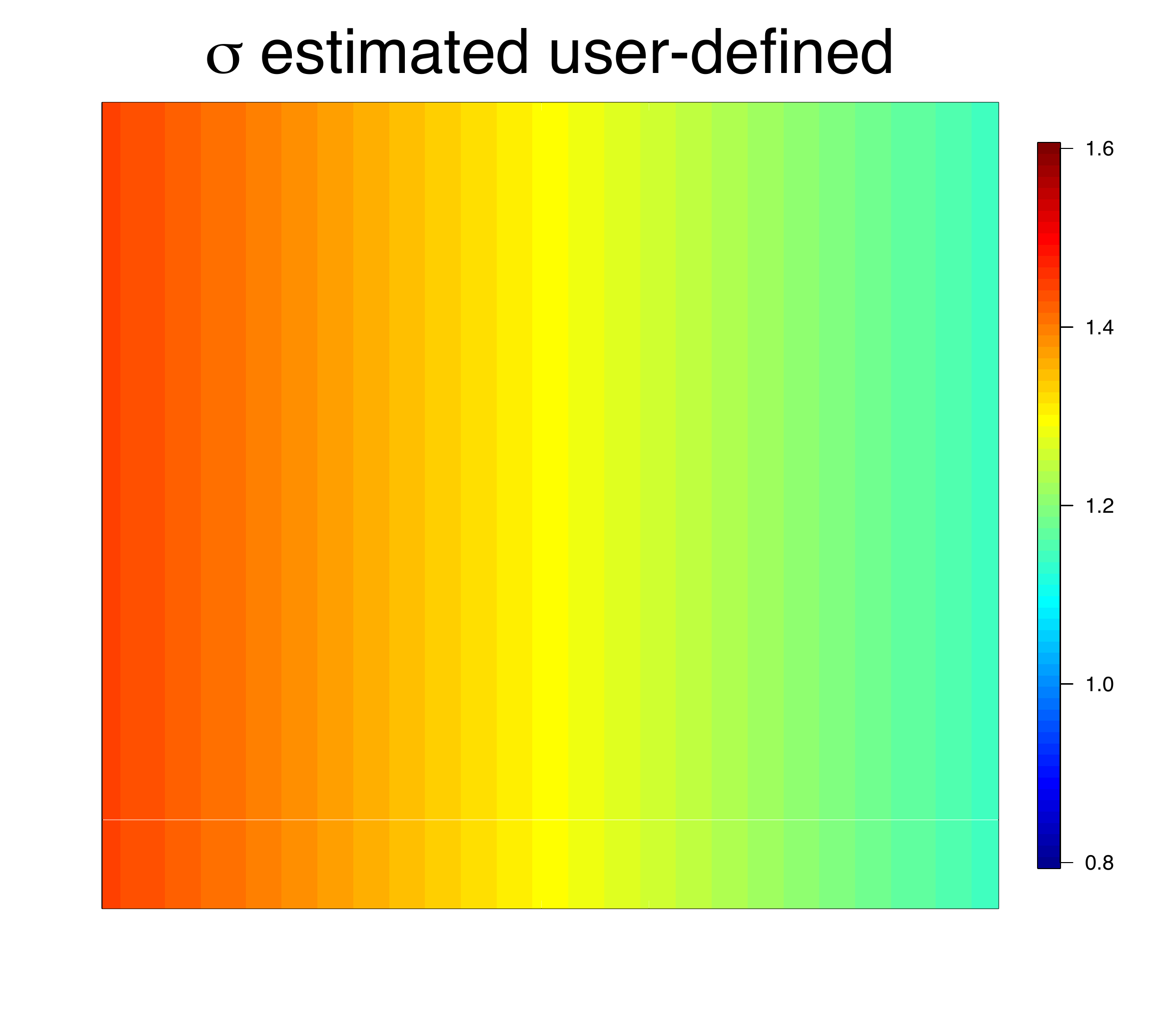}
\hspace{-2mm}
\includegraphics[width=0.32\textwidth]{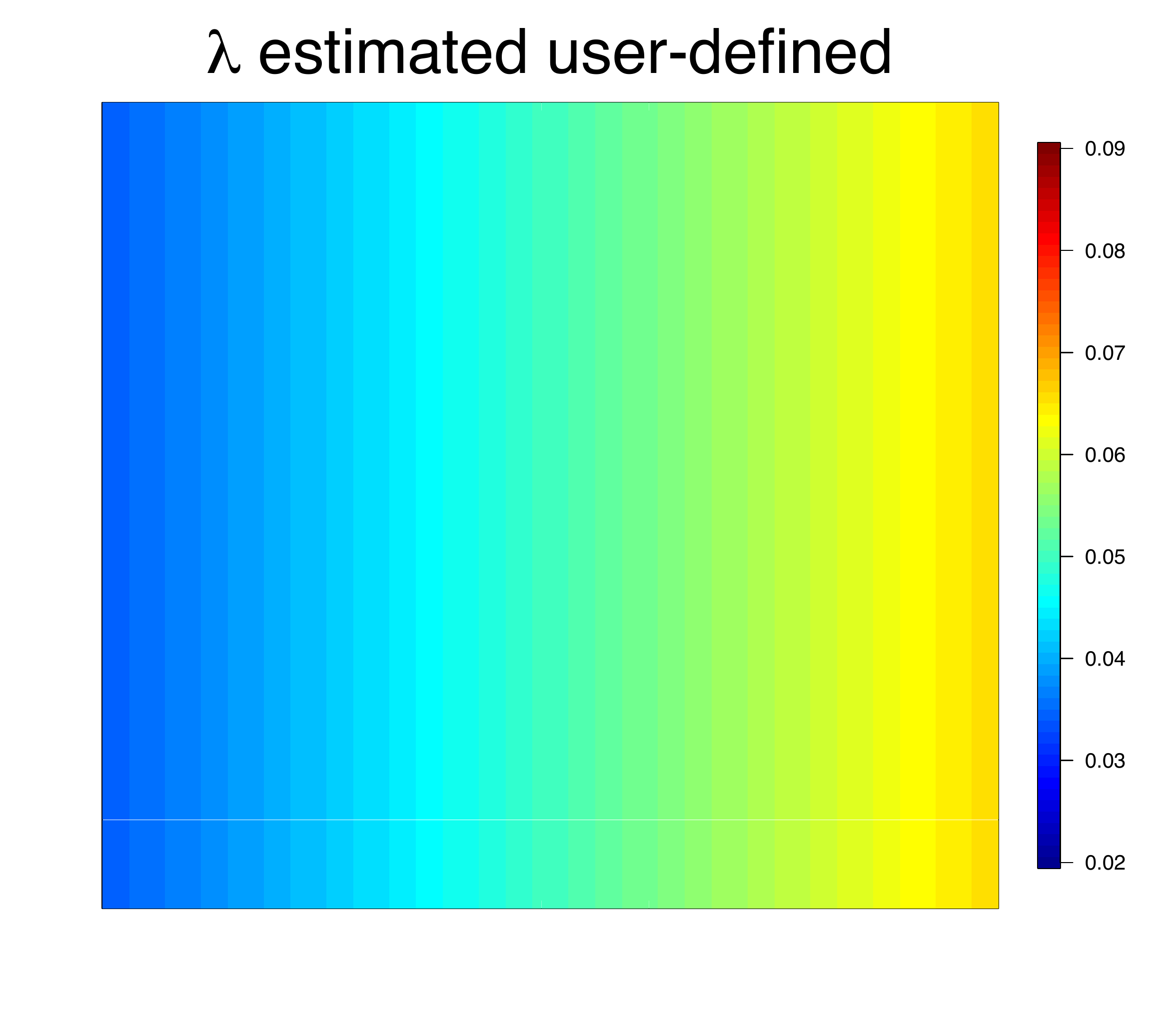}
\hspace{-2mm}
\includegraphics[width=0.32\textwidth]{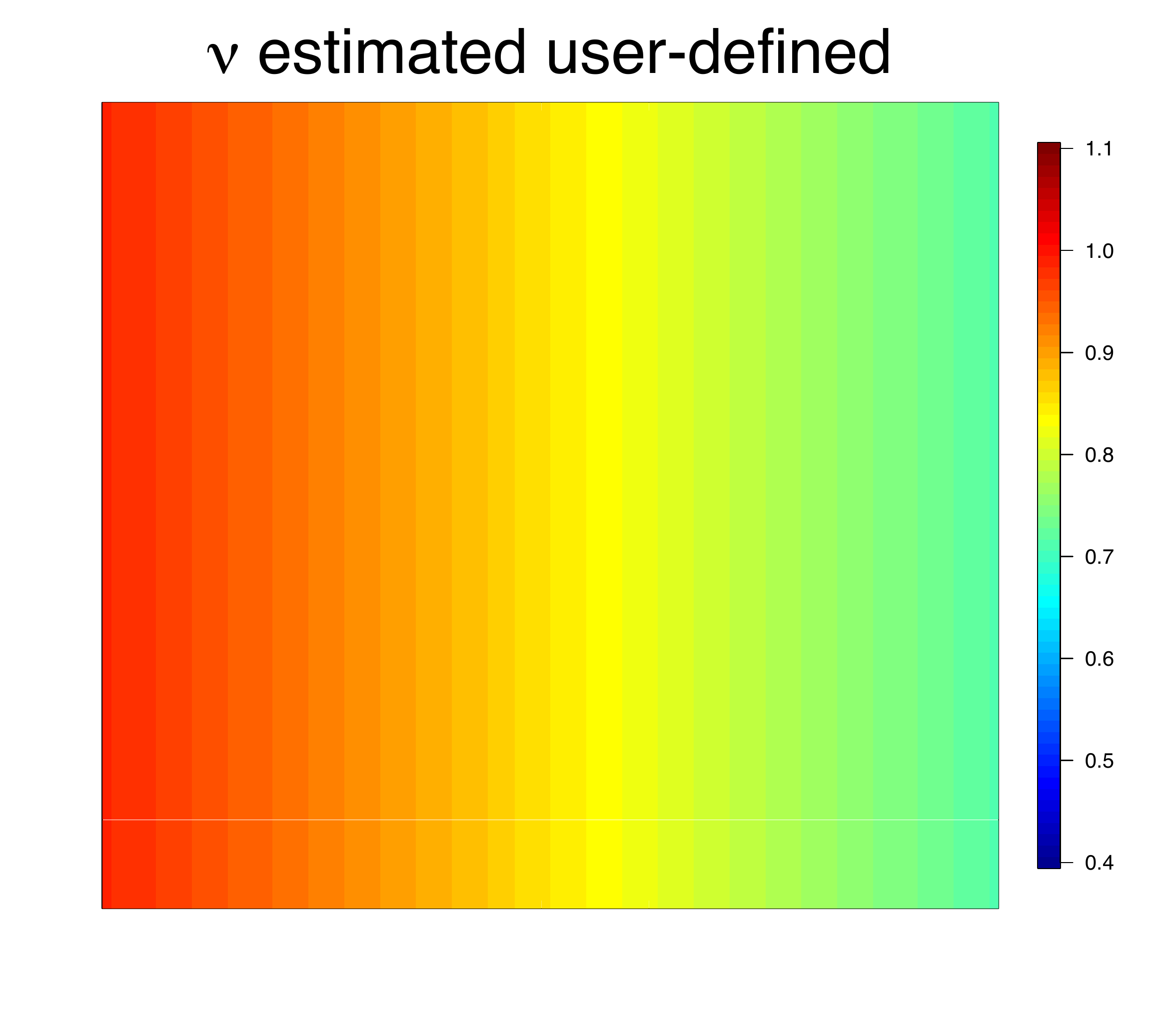}
\vspace{-4mm}
\caption{Estimated with two user-defined subregions. } 
\vspace{4mm}
 \end{subfigure}
  \hspace{8mm}
\begin{subfigure}[b]{0.46\textwidth}
\centering
\includegraphics[width=0.32\textwidth]{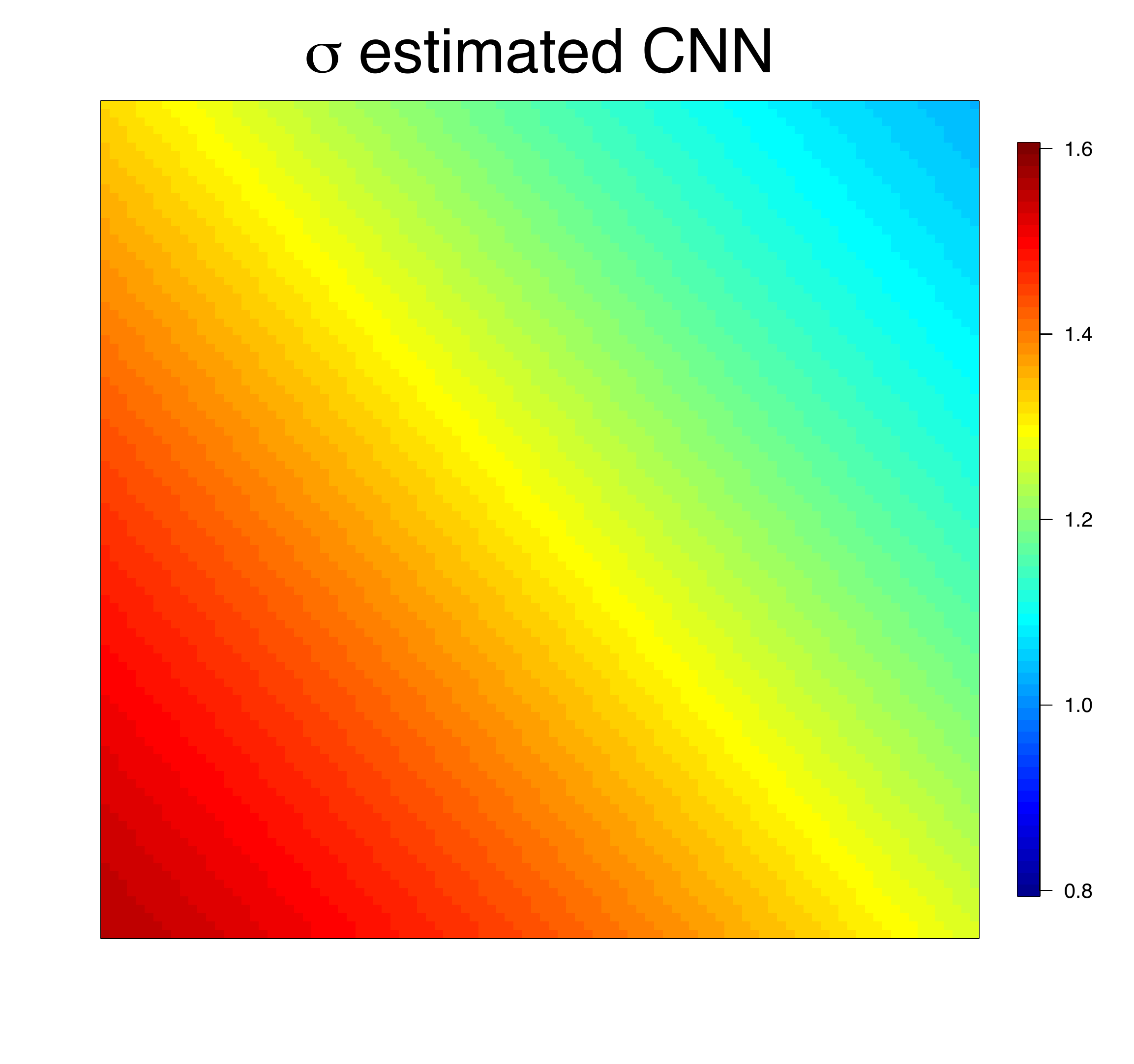}
\hspace{-2mm}
\includegraphics[width=0.32\textwidth]{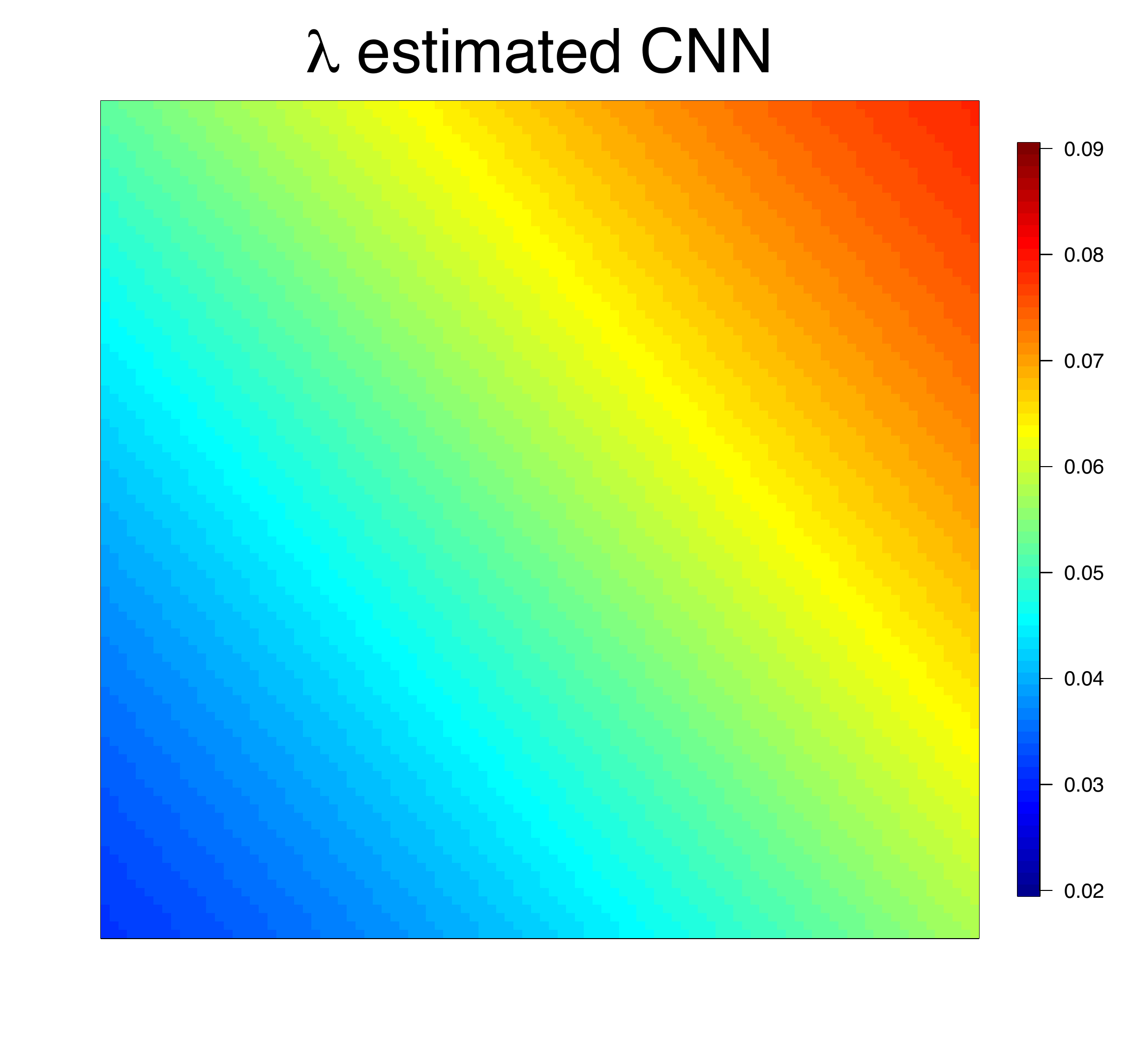}
\hspace{-2mm}
\includegraphics[width=0.32\textwidth]{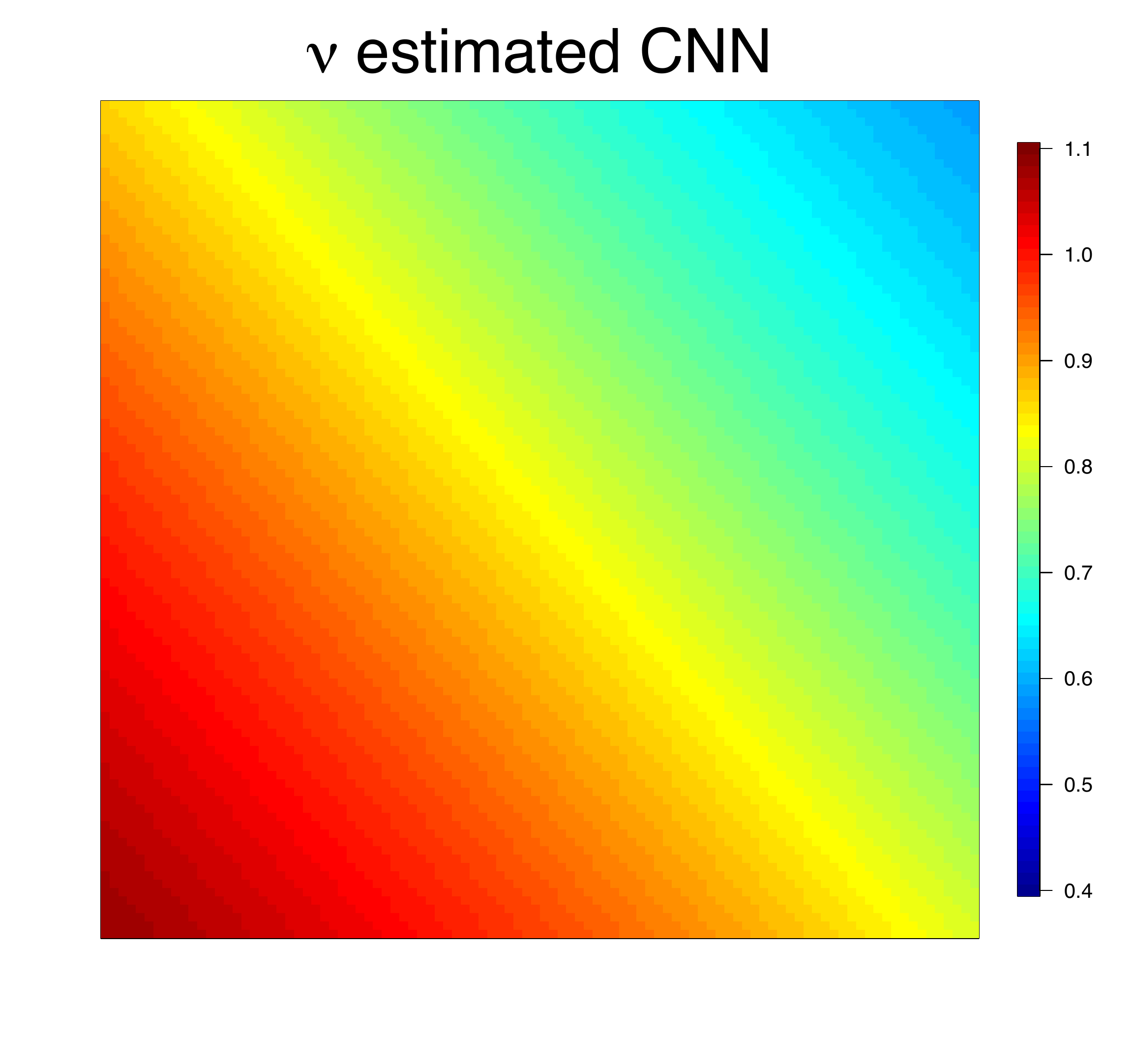}
\vspace{-4mm}
\caption{Estimated with two ConvNet subregions.} 
\vspace{4mm}
 \end{subfigure}
\end{minipage}
 \caption{Setting 1: Parameters generated with 4 subregions: The first row is the true parameter heatmap, the second row is the average of the estimated parameters with three subregions for user-defined as well as ConvNet-based subregions and the third row is the average of the estimated parameters with two subregions for user-defined as well as ConvNet-based subregions. 
 Setting 2: Parameters generated with 2 subregions: First row is the true parameter heatmap, the second row shows the average of the estimated parameters  by user-defined subregions and ConvNet-based subregions respectively.}
 \label{fig:simulation}
\end{figure}

In the first simulation scenario (Setting 1), we generated data with four subregions using the nonstationary Matérn covariance function $C^{NS}(\cdot,\cdot;\boldsymbol{\theta}(\cdot))$. The datasets were simulated on a regular grid of 10,000 points in the spatial domain $[0,1]^2$. The spatially varying parameter regions were defined using kernel smoothing, as explained in equation \eqref{eq6}. We selected four anchor locations: $\mathbf{a}_1 = (0.25,0.25)^\top$, $\mathbf{a}_2 = (0.25,0.75)^\top$, $\mathbf{a}_3 = (0.75,0.25)^\top$, and $\mathbf{a}_4 = (0.75,0.75)^\top$, with corresponding parameters: $\sigma_1 = 1.2$, $\lambda_1 = 0.05$, $\nu_1 = 0.9$, $\sigma_2 = 0.8$, $\lambda_2 = 0.02$, $\nu_2 = 0.4$, $\sigma_3 = 0.8$, $\lambda_3 = 0.02$, $\nu_3 = 0.4$, $\sigma_4 = 0.8$, $\lambda_4 = 0.02$, and $\nu_4 = 0.4$. The constant $\phi$ was set to be $\pi/2$ for all simulation scenarios. In this scenario, we estimated the parameters using two and three subregions, employing user-defined and ConvNet-based partitioning. The user-defined approach divided the regions equally into two or three subregions by splitting along the x-axis, with the anchor points chosen as the midpoints of these regions. On the other hand, the anchor points for ConvNet-based subregions were determined using Algorithm \ref{alg:alg1}, as discussed in Section \ref{Sec:clustering}. Both approaches have been implemented in the \pkg{ExaGeoStat} package and can be utilized to obtain parameter estimates based on exact likelihood computations. Compared to the true values, the average mean squared error (MSE) of spatially varying parameter estimates is presented in Table~\ref{tab:1}. Different methods and predefined numbers of subregions were employed for this analysis. The table demonstrates that ConvNet subregions yield smaller MSE and SE values than user-defined approaches. Furthermore, utilizing a number of subregions close to the true number of subregions reduces estimation errors. Additionally, Figure \ref{fig:simulation} in Setting 1 presents the heatmap of the parameter estimates averaged over all simulations. The results clearly indicate that the three ConvNet-based subregions outperforms the user-defined approach and accurately captures the true patterns of the parameters. It is noteworthy that even when the number of subregions was miss-specified, the estimated parameters were still able to capture the true parameter patterns effectively.

 


In the second simulation scenario (Setting 2), we follow the same formulation as the first one. However, this time we generate data with two subregions and estimate the parameters using two subregions using ConvNet subregions and user-defined subregions methods. This scenario focuses on comparing the efficiency of Algorithm \ref{alg:alg1} with the user-defined splitting algorithm. We select two anchor points for this setting: $\mathbf{a}_1 = (0.25,0.25)^\top$ and $\mathbf{a}_2 = (0.75,0.75)^\top$. The corresponding parameter values are $\sigma_1 = 1.6$, $\lambda_1 = 0.09$, $\nu_1 = 1.1$, $\sigma_2 = 0.8$, $\lambda_2 = 0.02$, and $\nu_2 = 0.4$. Using the user-defined subregion approach, we split the entire region into two halves along the x-axis. Table \ref{tab:1} shows that the parameter estimates are superior when using the ConvNet subregions compared to the user-defined splitting approach. This can also be observed in Figure \ref{fig:simulation}, where it is clear that the user-defined subregions led to the incorrect selection of anchor points, causing the model to fail in capturing the true spatial pattern of the parameters. However, the ConvNet-based algorithm adaptively selected the subregions, resulting in better parameter estimation.

\begin{table}[!htb]
    \centering
    \caption{Average Mean Square Error (MSE) and Standard Error (SE) for parameter estimates in different simulation settings.}
    \resizebox{\columnwidth}{!}{%
        \begin{tabular}{||c | c c c c c c c c||} 
        \hline
         & Method & Number of   &$MSE_{\sigma} $ & $SE_{\sigma}$ &$MSE_{\lambda}$ & $SE_{\lambda} $ & $MSE_{\nu} $ & $SE_{\nu} $ \\
         &  &  subregions &($\times 10^{-4}$) & ($\times 10^{-3}$) & ($\times 10^{-5}$) & ($\times 10^{-4})$ &
         ($\times 10^{-5}$) & ($\times 10^{-3}$)
         \\        
         \hline\hline

       
       Setting 1 & 
         \begin{tabular}{c} 
         ConvNet \\
         ConvNet\\
         User-defined\\
         User-defined \\
         \end{tabular} 
         & 
           \begin{tabular}{c} 
         3 \\
         2\\
         3\\
         2 \\
         \end{tabular} 
         &
         \begin{tabular}{c} 
         4.305 \\
         85.12\\
         36.86\\
         120.5 \\
         \end{tabular} 
         &
         \begin{tabular}{c} 
         2.461 \\
         3.97\\
         4.432 \\
         7.788 \\
         \end{tabular} 
         &
         
         \begin{tabular}{c} 
         3.322 \\
         4.171\\
         3.869\\
         15.72 \\
         \end{tabular} 
         &
         \begin{tabular}{c} 
         1.488 \\
         2.034\\
         1.733\\
         11.89 \\
         \end{tabular} 
         &
         
         \begin{tabular}{c} 
         1.084 \\
         866.2\\
         281.1\\
         1430.7 \\
         \end{tabular} 
         &
         
         \begin{tabular}{c} 
         3.511 \\
         3.721\\
         4.242\\
         17.39 \\
         \end{tabular} 
         \\ 
         
         \hline
         Setting 2 & 
         \begin{tabular}{c} 
         ConvNet \\
         user-defined\\ \end{tabular} 
         & 
           \begin{tabular}{c} 
         2 \\
         2\\ \end{tabular}
         &
         \begin{tabular}{c}
         234.3\\
         279.8\\\end{tabular} &
         \begin{tabular}{c}
         371.3\\ 
         439.2\\ \end{tabular} &
         
         \begin{tabular}{c} 
         4.689 \\ 
         7.078  \\ \end{tabular} &
         \begin{tabular}{c} 
         1.049\\ 
         2.549 \\ \end{tabular} &
         
         \begin{tabular}{c} 
         18.51\\ 
         612.3 \\ \end{tabular} &
         
         \begin{tabular}{c} 
         1.637\\ 
         2.835 \\ \end{tabular} \\ 
         \hline
         \end{tabular}%
         }
    \label{tab:1}
\end{table}




In the third simulation setting (Setting 3), we generated constant parameter values to assess whether our implementation could capture stationarity in parameters. We estimated the constant parameters using three subregions as well as four subregions, each with 100 replicates. Table \ref{tab:cons_param} presents the parameter estimation results, averaged over all subregions and replicates. It is evident that, on average, the model can estimate the parameter values very close to the true values. Additionally, the parameter estimates from each subregion are close to each other, indicating that the model can efficiently estimate stationary parameter spaces.
 
 \begin{table}[!htb]
    \centering
    \caption{Comparison of the average of the estimated parameters with the true parameter settings for simulation Setting 3.}
    \begin{tabular}{||c | c c c||} 
    \hline
    Parameters & $\sigma$ & $\lambda$ & $\nu$ \\ [0.5ex] 
    \hline\hline
    True parameters & 2 & 0.15 & 0.8 \\
    \hline
     Three ConvNet subregions & 
     \begin{tabular}{c} 
         2.235\\
         2.077\\ 
         2.891\\ \end{tabular} & 
         \begin{tabular}{c}
         0.137 \\ 
         0.159 \\ 
         0.164 \\ \end{tabular} &
         \begin{tabular}{c} 
         0.813\\ 
         0.798\\ 
         0.814 \\ \end{tabular}
         \\ 
    \hline
     Four ConvNet subregions & 
     \begin{tabular}{c} 
         1.983\\
         1.837\\ 
         2.391\\ 
         1.939\\ \end{tabular} & 
         \begin{tabular}{c}
         0.172 \\ 
         0.155 \\ 
         0.178 \\ 
         0.128 \\ \end{tabular} &
         \begin{tabular}{c} 
         0.821\\ 
         0.796\\ 
         0.761 \\ 
         0.793 \\ \end{tabular}
         \\ 
    \hline
    \end{tabular}
    
    \label{tab:cons_param}
\end{table}

\section{Soil Moisture Data Application} \label{Sec:real_data}

This section focuses on applying our method to analyze soil moisture content data across the Mississippi Basin region in the United States. The data used in this study were sourced from the HydroBlocks numerical land surface model \citep{chaney2016hydroblocks}. The study area encompasses a large geographic extent of approximately 2,400,000 square kilometers, spanning longitudes from $-$107.7166 to $-$92.47494 and latitudes from 32.37106 to 43.43772. Due to the varied topography within the region, the soil moisture content exhibits nonstationarity. \cite{huang2018hierarchical} analyzed this dataset with a 1-kilometer resolution. They first removed a spatial mean function and then fitted a stationary Gaussian process to the transformed residuals. In this study, we conduct the same exploratory data analysis but only consider a coarser resolution of 10 kilometers for those residuals and fit the proposed nonstationary Gaussian process. In addition, our analysis focuses on a subset of $200{,}000$ locations selected explicitly for training purposes. As a preprocessing step, we apply a zero-mean Gaussian process model with a Matérn covariance function.

We employ a mean-zero Gaussian likelihood with a nonstationary Mat\'ern covariance, incorporating three spatially varying parameters ${\sigma(\mathbf{s}),\lambda(\mathbf{s}),\nu(\mathbf{s})}$, and set $\phi = \pi/2$. We assess the parameter estimations using three and four ConvNet subregions, respectively. For these extensive computations, we leverage the Shaheen-II supercomputing facility at KAUST. Shaheen-II is a Cray XC40 system with $6{,}174$ dual-socket compute nodes based on 16-core Intel Haswell processors running at 2.3 GHz, where each node has 128 GB of DDR4 memory. The Shaheen-II system has $197{,}568$ processor cores and 790 TB of aggregate memory. We used $256$ nodes to do our experiments on the real dataset. The MLE optimization process lasted approximately 17.53 hours and 21.62 hours for the three-cluster and four-cluster configurations, respectively.

The parameter estimates obtained using the nonstationary Matérn covariance model applied to the soil moisture data are depicted in Figure 5. The divisions in each subregion are visibly linear rather than non-linear. This linearity arises from the fact that the subregions are formed using a minimum distance criterion from the node location. Since the dataset is structured on a regular grid, a linear border based on distance will always be generated. To evaluate the models' performance, we also computed the Akaike Information Criterion (AIC). AIC for three subregion was $-274390.7$ and for four subregions it was -273408.3. It is evident from the findings that the three subregion modeling framework produces the most accurate parameter estimations for the soil moisture data.

\begin{figure}[htp!]
\captionsetup[subfigure]{labelformat=empty}
   \begin{minipage}[t]{\linewidth}
       \vspace{3mm}
   \subcaption{Soil Moisture Data}
 \vspace{1mm}
 \begin{subfigure}[b]{\textwidth}
\centering
\includegraphics[width=0.45\textwidth]{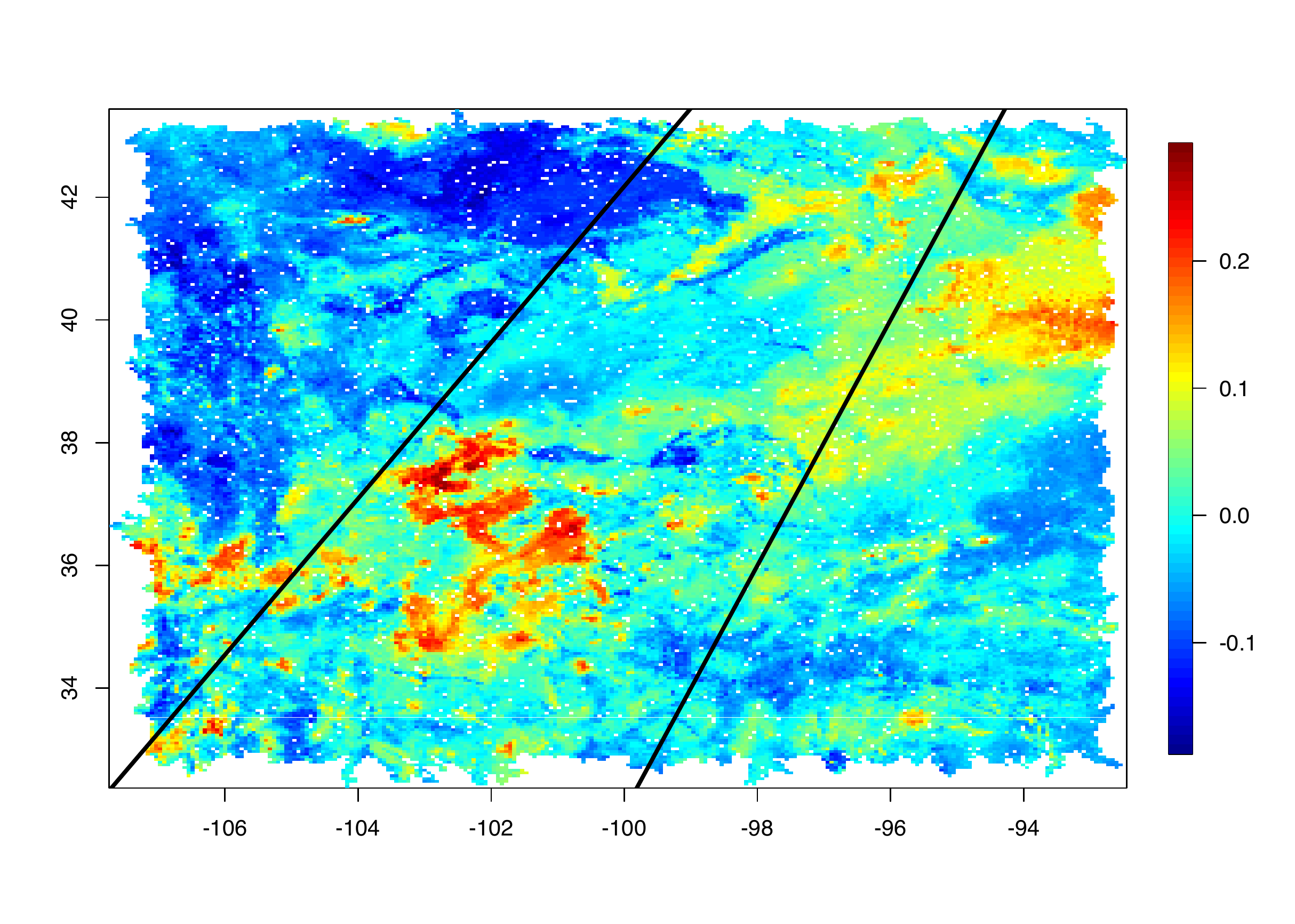}
\hspace{-2mm}
\includegraphics[width=0.45\textwidth]{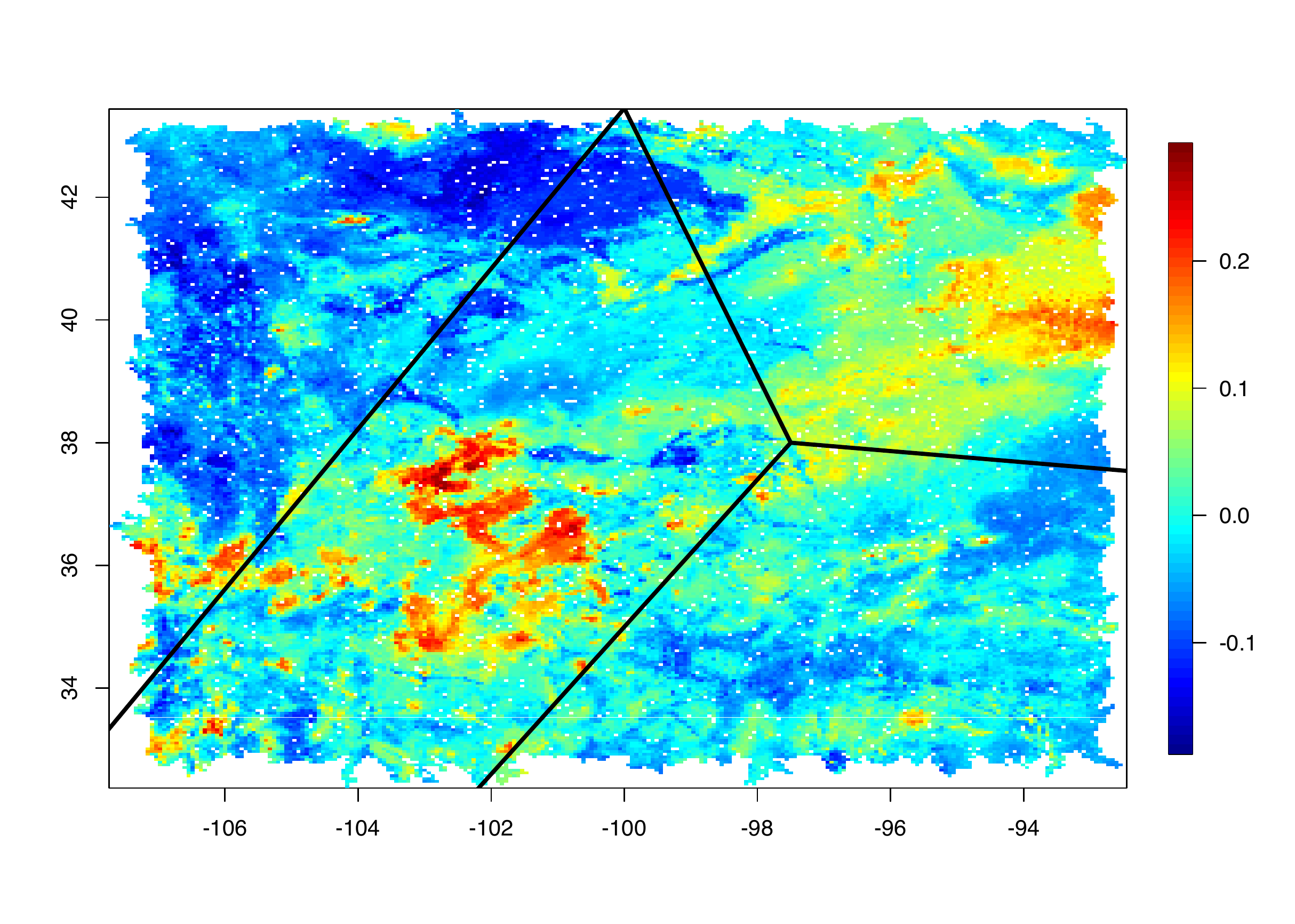}
\vspace{-4mm}
\caption{The black lines show the splits for three and four subregions, respectively.} 
\vspace{3mm}
 \end{subfigure}
\end{minipage}
   \begin{minipage}[t]{\linewidth}
       \vspace{3mm}
 \vspace{-2mm}
\centering
\begin{subfigure}[b]{\textwidth}
\centering
\includegraphics[width=0.32\textwidth]{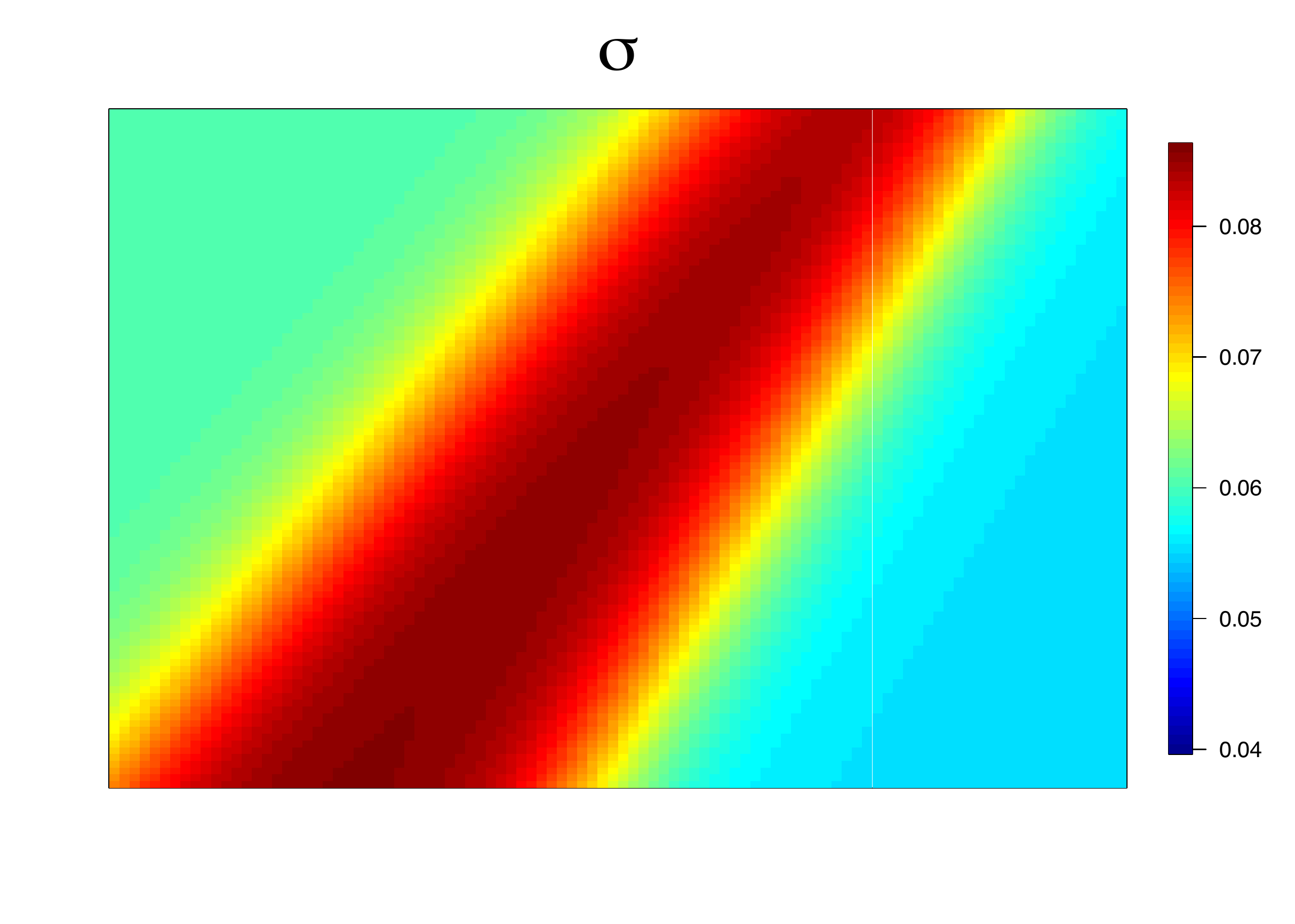}
\hspace{-2mm}
\includegraphics[width=0.32\textwidth]{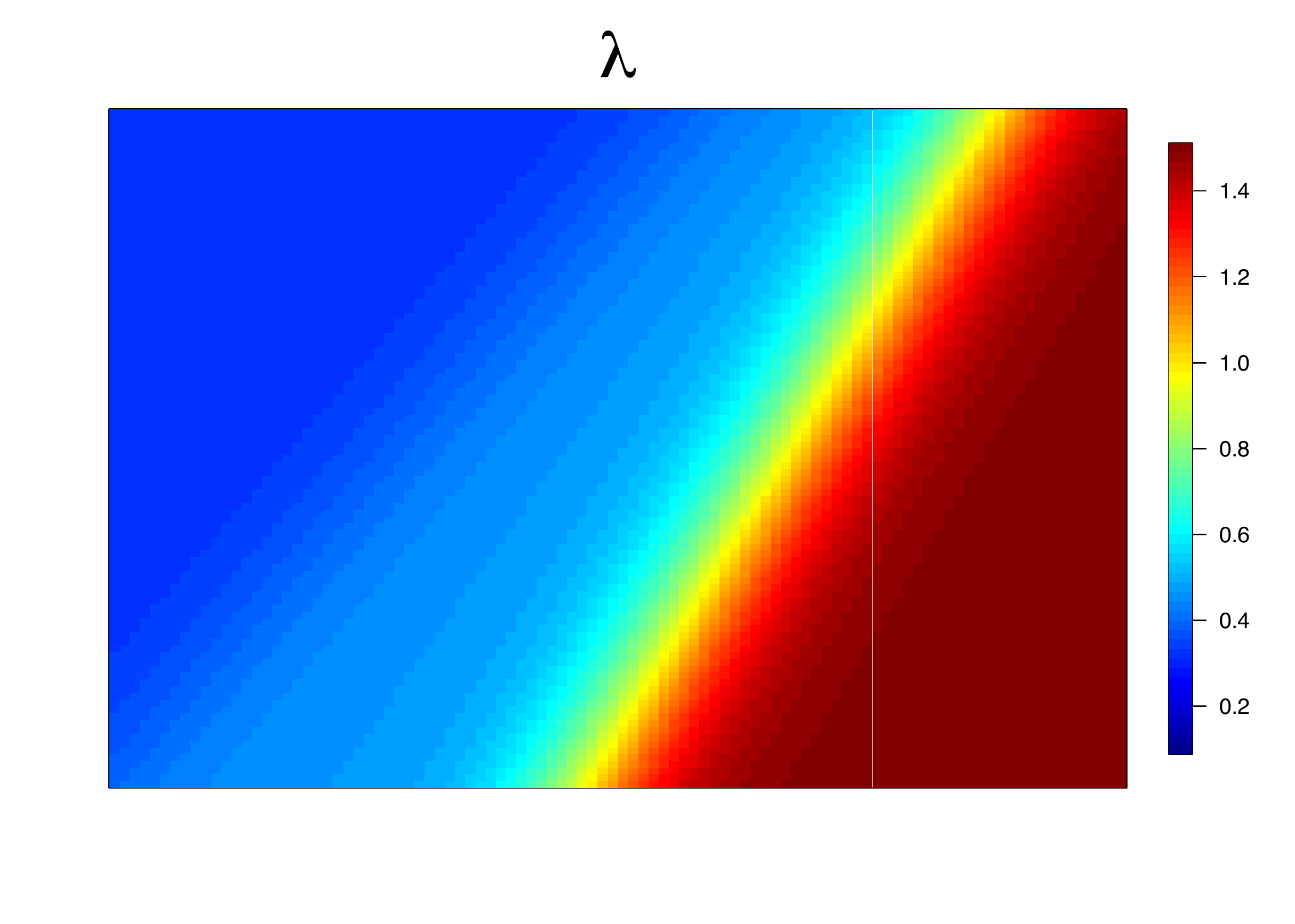}
\hspace{-2mm}
\includegraphics[width=0.32\textwidth]{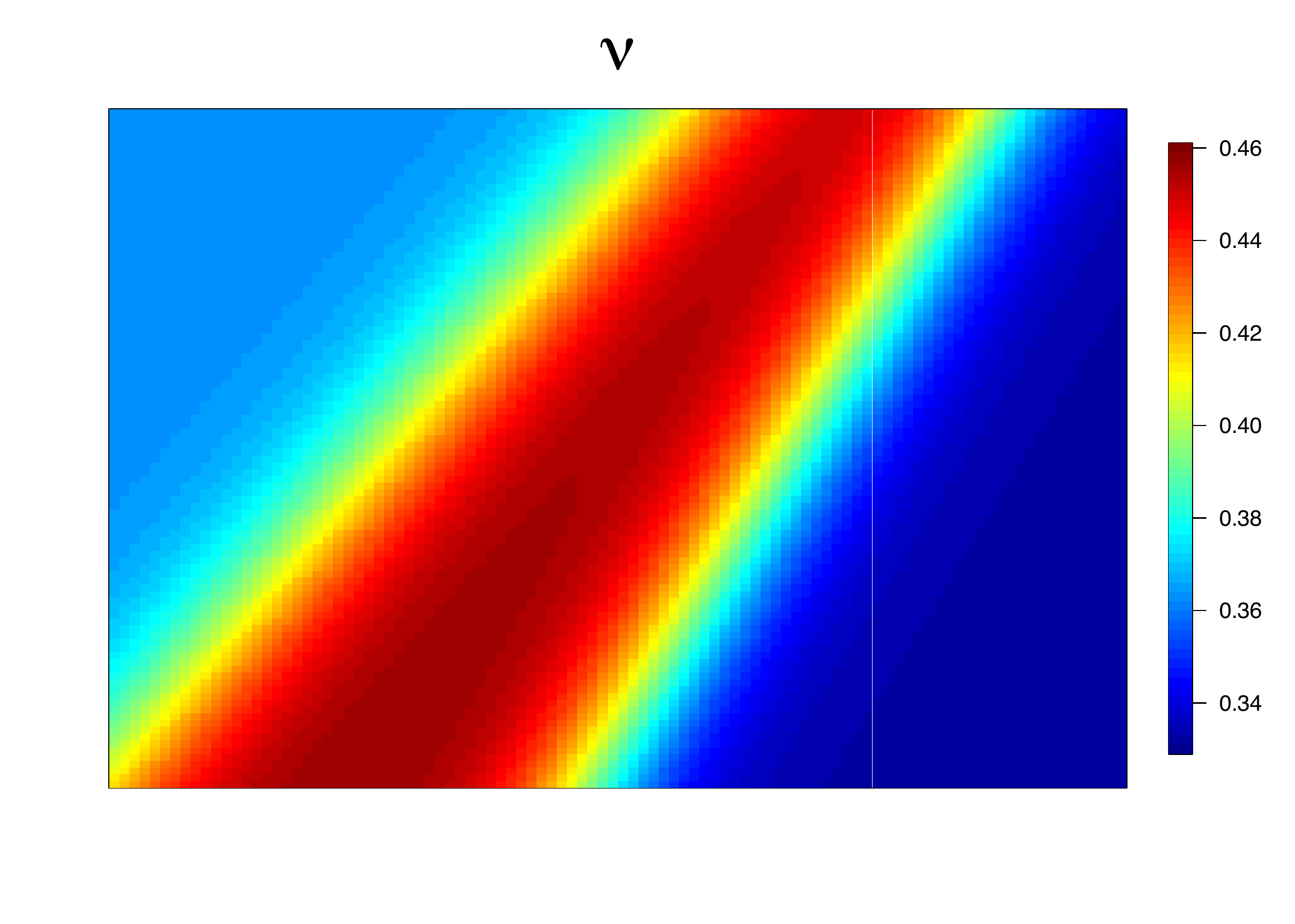}
\vspace{-4mm}
\caption{Estimated with three ConvNet subregions. } 
\vspace{4mm}
 \end{subfigure}
\end{minipage}

   \begin{minipage}[t]{\linewidth}
       \vspace{3mm}

 \vspace{-2mm}
\centering
\begin{subfigure}[b]{\textwidth}
\centering
\includegraphics[width=0.32\textwidth]{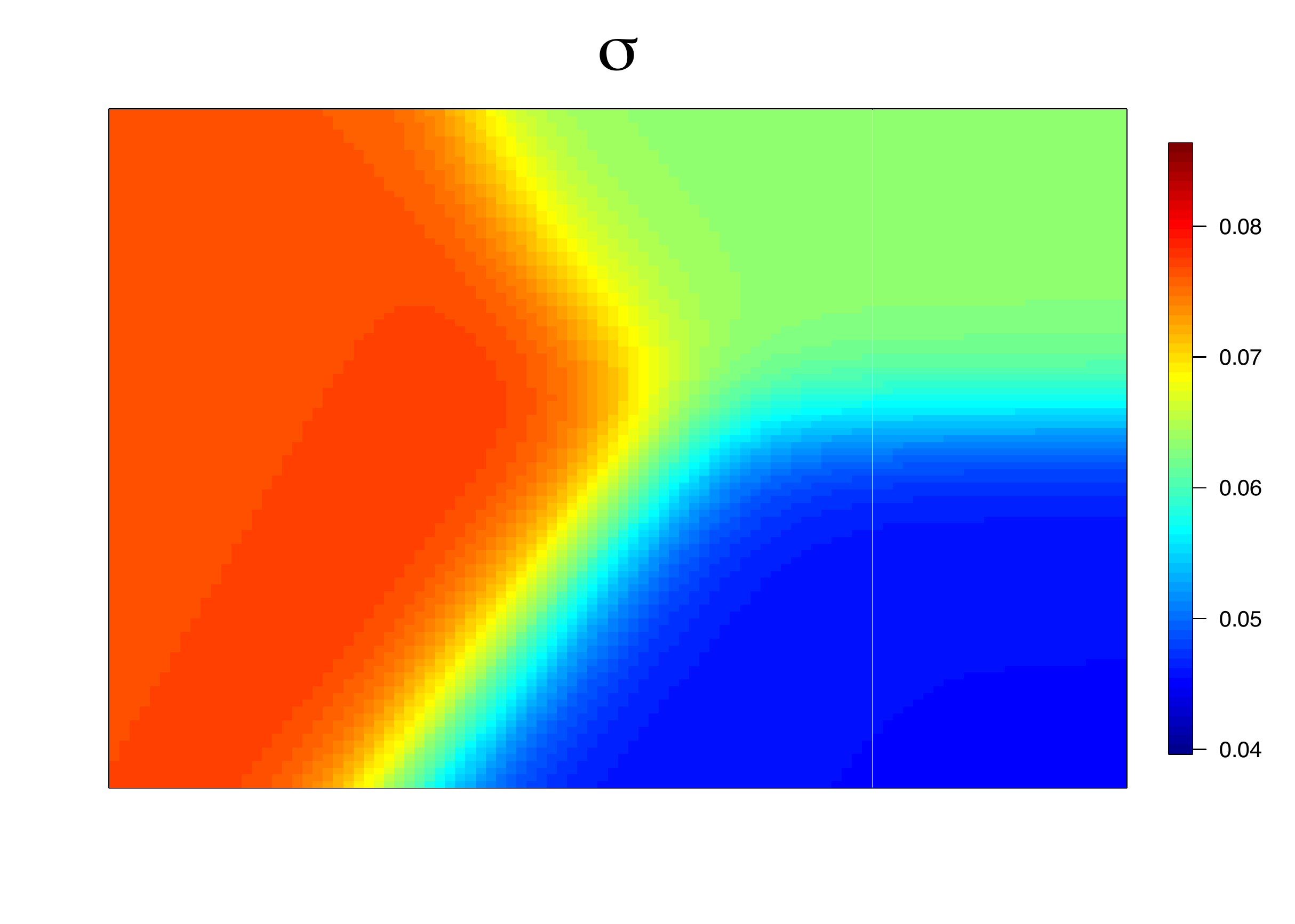}
\hspace{-2mm}
\includegraphics[width=0.32\textwidth]{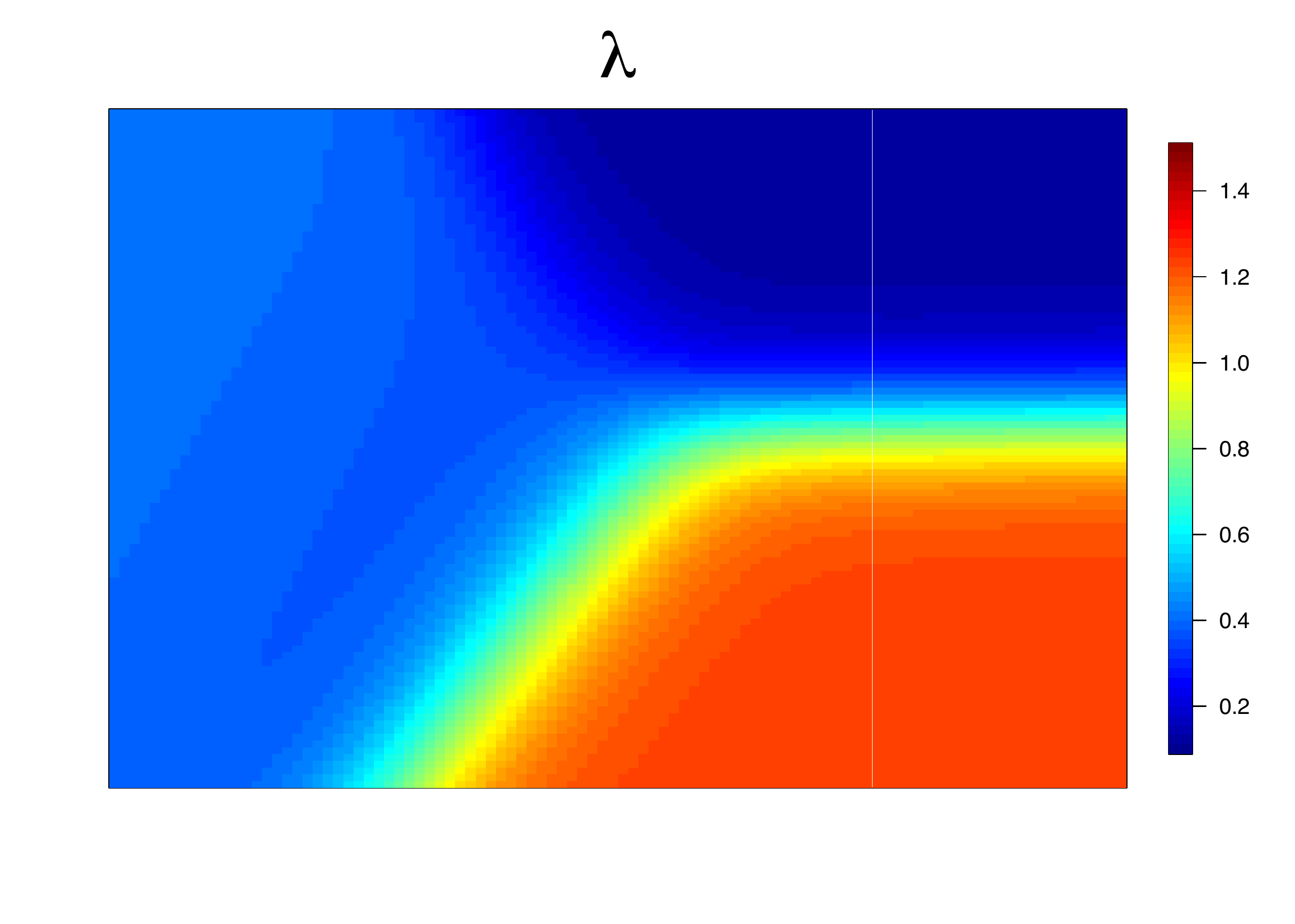}
\hspace{-2mm}
\includegraphics[width=0.32\textwidth]{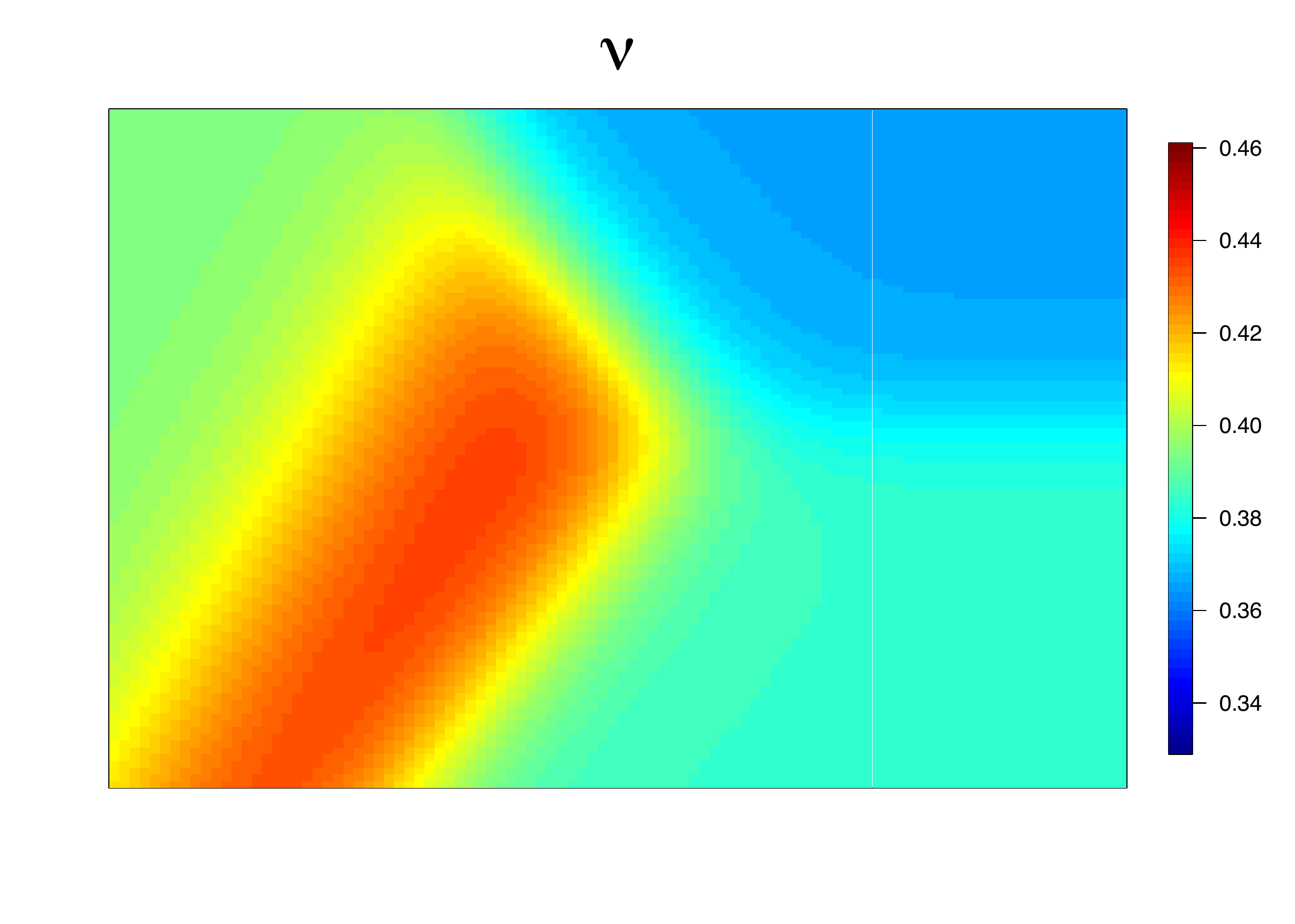}
\vspace{-4mm}
\caption{Estimated with four ConvNet subregions. } 
\vspace{4mm}
 \end{subfigure}
\end{minipage}
 \caption{Heatmap of the soil moisture data with its spatially varying parameter estimates over three and four subregion partitions respectively.}
 \label{fig:5}
\end{figure}

    

\section{Discussion} \label{Sec:discussion}

In this study, we presented a large-scale approach for modeling nonstationary phenomena in spatial data by utilizing dynamic partitioning with the help of ConvNet. Our framework offers a standalone solution that assigns a probability to each spatial region, enabling the differentiation between stationary and nonstationary processes. To evaluate the effectiveness of our framework, we conducted an individual assessment using both stationary and nonstationary datasets generated from the \pkg{ExaGeoStat} software. Furthermore, we integrated our ConvNet framework into \pkg{ExaGeoStat}, enabling dynamic partitioning of a given spatial region and selecting partitions with low stationary probabilities. This integration improves modeling capabilities under the assumption of nonstationarity.

Herein, we utilized kernel convolution as the primary method for generating parameter spaces when estimating nonstationary parameters. This approach distinguishes itself from traditional techniques such as weighted local stationary and local stationary methods. Instead of independently calculating the likelihood for each subregion, we adopted a ConvNet data-driven framework where we computed the likelihood for the entire dataset while selecting subregions. As a result, we can visualize the spatially varying parameter as a smooth function in space by estimating the parameters at representative node points. The proposed ConvNet framework has been trained specifically with Matérn stationary and nonstationary fields for Gaussian likelihood. However, it is important to note that this training only covers a limited range of potential spatial processes. Therefore, in future research, there is room for expansion by training the framework with various simulation types to distinguish between stationary and nonstationary processes even more effectively.

 
As part of future research, it is possible to replace the ConvNet model discussed in Section~\ref{Sec:ConvNet_model} with graph neural networks (GNNs) \citep{scarselli2008graph,zhou2020graph}. Real-world objects often have inherent connections to other entities, and these relationships can be naturally represented as a graph. GNNs, developed over a decade, are neural networks specifically designed to operate on graph data \citep{scarselli2008graph}. In recent years, GNNs have been successfully applied in various domains such as antibacterial discovery \citep{stokes2020deep}, physics simulations \citep{sanchez2020learning}, fake news detection \citep{monti2019fake}, traffic prediction \citep{nabi2023deep}, and recommendation systems \citep{eksombatchai2018pixie}. Notably, GNNs have also been used to model spatial processes, including deep Gaussian Markov random fields \citep{oskarsson2022scalable}.
In the context of this study, ConvNets can be easily extended to accommodate graphical structures \citep{quek2011structural}, enabling the classification of random processes using GNNs. This integration would contribute to developing a robust framework capable of capturing any spatial pattern, regardless of its regularity or irregularity. Additionally, this approach would reduce the necessity for extensive data pre-processing, as discussed in Section \ref{Sec:Data_Preprocessing}.

\section{Acknowledgement}
The funding for the research presented in this manuscript was provided by the King Abdullah University of Science and Technology (KAUST) in Thuwal, Saudi Arabia. We also would like to thank the Supercomputing Laboratory (KSL) at KAUST for providing computational resources to this project on the Shaheen-II Cray XC40 Supercomputer.

\bibliographystyle{Perfect}

\bibliography{Bibliography-MM-MC}
\end{document}